\documentclass[10pt,twocolumn,letterpaper]{article}

\usepackage{cvpr}              %

\usepackage{graphicx}
\usepackage{amsmath}
\usepackage{amssymb}
\usepackage{booktabs}
\usepackage{enumitem}
\usepackage[square,numbers,sort&compress]{natbib}
\usepackage{algpseudocode}
\usepackage[table]{xcolor}
\newcolumntype{x}[1]{>{\centering\arraybackslash}p{#1pt}}
\newcolumntype{y}[1]{>{\raggedright\arraybackslash}p{#1pt}}
\newcolumntype{z}[1]{>{\raggedleft\arraybackslash}p{#1pt}}
\newcolumntype{H}{>{\setbox0=\hbox\bgroup}c<{\egroup}@{}}
\newcommand{\tablestyle}[2]{\setlength{\tabcolsep}{#1}\renewcommand{\arraystretch}{#2}\centering\footnotesize}
\usepackage{array}
\usepackage{multirow}
\usepackage{algorithm}
\usepackage{subcaption}
\usepackage[normalem]{ulem}
\usepackage{xparse}
\usepackage{pifont}
\usepackage{bm}
\usepackage{threeparttable}
\usepackage{etoolbox}
\usepackage{listings}
\usepackage{mwe} %
\usepackage{makecell}
\usepackage{color, colortbl}
\usepackage{tabularx}
\usepackage[accsupp]{axessibility}
\usepackage{comment}
\usepackage{soul}
\makeatletter
\@namedef{ver@everyshi.sty}{}
\makeatother
\usepackage{tikz}
\usepackage{dblfloatfix}
\usepackage{pgfplots}
\pgfplotsset{compat=1.16}
\usepgfplotslibrary{groupplots}
\usetikzlibrary{patterns}
\usepackage[T1]{fontenc}
\usepackage{tablefootnote}

\usepackage{pgfkeys}
	\newenvironment{customlegend}[1][]{%
		\begingroup
		\csname pgfplots@init@cleared@structures\endcsname
		\pgfplotsset{#1}%
	}{%
		\csname pgfplots@createlegend\endcsname
		\endgroup
	}%
	\def\addlegendimage{\csname pgfplots@addlegendimage\endcsname}

\usetikzlibrary{arrows}
\usepackage{tkz-kiviat}
\usepackage{xfp}

\usepackage{pifont}
\newcommand{\cmark}{\text{\ding{51}}}
\newcommand{\xmark}{\text{\ding{55}}}

\newcommand{\vu}{\mathbf{u}}
\newcommand{\vv}{\mathbf{v}}

\newcommand{\calX}{\mathcal{X}}
\newcommand{\calY}{\mathcal{Y}}
\DeclareRobustCommand{\hlGray}[1]{{\sethlcolor{Gray}\hl{#1}}}

\newcommand{\thickhline}{\Xhline{3\arrayrulewidth}}

\definecolor{citecolor}{HTML}{0071bc}
\definecolor{color_ao}{gray}{0.5}
\definecolor{color_our}{rgb}{0.66,0.82,0.56}
\definecolor{color_pre}{rgb}{0.52,0.59,0.69}
\definecolor{Gray}{gray}{0.9}
\definecolor{LighterGray}{gray}{0.93}
\definecolor{LightGrayForTableRule}{gray}{0.92}
\definecolor{DarkGray}{gray}{0.5}
\definecolor{Black}{rgb}{0.0, 0.0, 0.0}
\definecolor{NiceBlue}{rgb}{0.11764705882352941, 0.5647058823529412, 1.0}
\definecolor{NiceGreen}{rgb}{0.0, 0.5, 0.0}
\usepackage[breaklinks=true,bookmarks=false,colorlinks,bookmarks=false, citecolor=citecolor]{hyperref}

\usepackage[capitalize]{cleveref}
\crefname{section}{\S}{\S\S}
\crefname{subsection}{\S}{\S\S}
\crefformat{table}{Table~#2#1#3}
\crefformat{figure}{Figure~#2#1#3}
\crefformat{equation}{Eq~#2#1#3}

\newcommand{\myparagraph}[1]{\vspace{0pt}\noindent{\bf #1}}
\newcommand{\ours}{\textsc{LaViLa}\xspace}
\newcommand{\narrator}{\textsc{Narrator}\xspace}
\newcommand{\rephraser}{\textsc{Rephraser}\xspace}

\newcommand{\ek}{EK-100\xspace}
\newcommand{\ekmir}{EK-100 MIR\xspace}
\newcommand{\ekcls}{EK-100 CLS\xspace}
\newcommand{\egomcq}{EgoMCQ\xspace}
\newcommand{\egonlq}{EgoNLQ\xspace}
\newcommand{\charadesego}{CharadesEgo\xspace}

\begin{document}

\title{Learning Video Representations from Large Language Models}

\author{
	Yue Zhao$^{1,2}$\thanks{Work done during an internship at Meta.} \quad Ishan Misra$^{1}$ \quad Philipp Kr\"ahenb\"uhl$^{2}$ \quad Rohit Girdhar$^{1}$ \\
	$^{1}$FAIR, Meta AI \quad $^{2}$University of Texas, Austin \\
	{\small \href{https://facebookresearch.github.io/LaViLa}{\tt facebookresearch.github.io/LaViLa}}
}

\maketitle

\begin{abstract}
  We introduce {\bf \ours}, a new approach to learning video-language representations by leveraging Large Language Models (LLMs). We repurpose pre-trained LLMs to be conditioned on visual input, and finetune them to create automatic video narrators. Our auto-generated narrations offer a number of advantages, including dense coverage of long videos, better temporal synchronization of the visual information and text, and much higher diversity of text. The video-text embedding learned contrastively with these additional auto-generated narrations outperforms the previous state-of-the-art on multiple first-person and third-person video tasks, both in zero-shot and finetuned setups. Most notably, \ours obtains an absolute gain of {\bf 10.1\%} on EGTEA classification and {\bf 5.9\%} Epic-Kitchens-100 multi-instance retrieval benchmarks. Furthermore, \ours trained with only half the narrations from the Ego4D dataset outperforms baseline models trained on the full set, and shows positive scaling behavior on increasing pre-training data and model size.
\end{abstract}
\section{Introduction}

\begin{figure}
 	\vspace{-5pt}
	\resizebox{0.9\linewidth}{!}{
		\newcommand{\lattice}{4}
\newcommand{\naxis}{8}
\newcommand{\amax}{66.5}
\newcommand{\amin}{59.4}

\newcommand{\bmax}{50.9}
\newcommand{\bmin}{45.0}

\newcommand{\cmax}{36.1}
\newcommand{\cmin}{32.1}

\newcommand{\dmax}{63.1}
\newcommand{\dmin}{57.2}

\newcommand{\emax}{76.0}
\newcommand{\emin}{65.9}

\newcommand{\fmax}{51.0}
\newcommand{\fmin}{50.5}

\newcommand{\gmax}{88.1}
\newcommand{\gmin}{82.7}

\newcommand{\hmax}{61.5}
\newcommand{\hmin}{54.3}

\newcommand{\origin}{0.8}

\newcommand\ColorBox[1]{\textcolor{#1}{\rule{3ex}{3ex}}}

\newcommand{\annotMark}[5]{
	\pgfmathsetmacro{\xcor}{#3*cos{(#1*#2)}/(1/#4)};
	\pgfmathsetmacro{\ycor}{#3*sin{(#1*#2)}/(1/#4)};
	\draw (\xcor,\ycor)node[anchor=south]{#5};
}

\begin{tikzpicture}[rotate=0, scale=0.95,every node/.style={inner sep=-15,outer sep=-15}]
	\tkzKiviatDiagram[lattice=\lattice, gap=1, step=1, label space=1.6]
	{EK-100 Multi-Instance Retrieval (nDCG),
		EK-100 Multi-Instance Retrieval (mAP),
		CharadesEgo Recognition \\(mAP),
		EgoMCQ \\(intra-vid. acc.),
		EGTEA Recognition (mean acc.),
		EK-100 Recognition (top-1 acc.),
		UCF-101 Recognition (linear probing mean acc.),
		HMDB-51 Recognition (linear probing mean acc.)}
	
		\tkzKiviatLine[thick, fill=color_our!80, color=NiceGreen, opacity=0.5](
		\fpeval{(66.5/\amax-\origin)/(1 - \origin)*\lattice},
		\fpeval{(50.9/\bmax-\origin)/(1 - \origin)*\lattice},
		\fpeval{(36.1/\cmax-\origin)/(1 - \origin)*\lattice},
		\fpeval{(63.1/\dmax-\origin)/(1 - \origin)*\lattice},
		\fpeval{(76.0/\emax-\origin)/(1 - \origin)*\lattice},
		\fpeval{(51.0/\fmax-\origin)/(1 - \origin)*\lattice},
		\fpeval{(88.1/\gmax-\origin)/(1 - \origin)*\lattice},
		\fpeval{(61.5/\hmax-\origin)/(1 - \origin)*\lattice})
		\tkzKiviatLine[thick, fill=gray!50, color=gray, opacity=0.5](
		\fpeval{(59.4/\amax-\origin)/(1 - \origin)*\lattice},
		\fpeval{(45.0/\bmax-\origin)/(1 - \origin)*\lattice},
		\fpeval{(32.1/\cmax-\origin)/(1 - \origin)*\lattice},
		\fpeval{(57.2/\dmax-\origin)/(1 - \origin)*\lattice},
		\fpeval{(65.9/\emax-\origin)/(1 - \origin)*\lattice},
		\fpeval{(50.5/\fmax-\origin)/(1 - \origin)*\lattice},
		\fpeval{(82.7/\gmax-\origin)/(1 - \origin)*\lattice},
		\fpeval{(54.3/\hmax-\origin)/(1 - \origin)*\lattice})
	\annotMark{0}{360/\naxis}{2}{1}{\amin};
	\annotMark{0}{360/\naxis}{3.4}{1}{\amax};
	\annotMark{1}{360/\naxis}{2.5}{1}{\bmin};
	\annotMark{1}{360/\naxis}{3.8}{1}{\bmax};
	\annotMark{2}{360/\naxis}{2.7}{1}{\cmin};
	\annotMark{2}{360/\naxis}{4.2}{1}{\cmax};
	\annotMark{3}{360/\naxis}{2.7}{1}{\dmin};
	\annotMark{3}{360/\naxis}{3.8}{1}{\dmax};
	\annotMark{4}{360/\naxis}{1.9}{1}{\emin};
	\annotMark{4}{360/\naxis}{3.4}{1}{\emax};
	\annotMark{5}{360/\naxis}{2.8}{1}{\fmin};
	\annotMark{5}{360/\naxis}{3.7}{1}{\fmax};
	\annotMark{6}{360/\naxis}{1.7}{1}{\gmin};
	\annotMark{6}{360/\naxis}{3.7}{1}{\gmax};
	\annotMark{7}{360/\naxis}{2}{1}{\hmin};
	\annotMark{7}{360/\naxis}{3.2}{1}{\hmax};
	\node[anchor=south west,xshift=-60pt,yshift=20pt] at (current bounding box.south east)
{
	\begin{tabular}{@{}lp{3cm}@{}}
		\ColorBox{color_our!80} & \ours (Ours) \\
		\ColorBox{gray!50} & Previous SOTA \\
	\end{tabular}
};
\end{tikzpicture}%
	}
	\caption{
		{\bf \ours sets a new state-of-the-art} across a number of first and third-person video understanding tasks (\cf~\cref{tab:datasets} for details), by learning a video-language representation using supervision from large language models as narrators.
	}
	\label{fig:teaser_sota}
\end{figure}
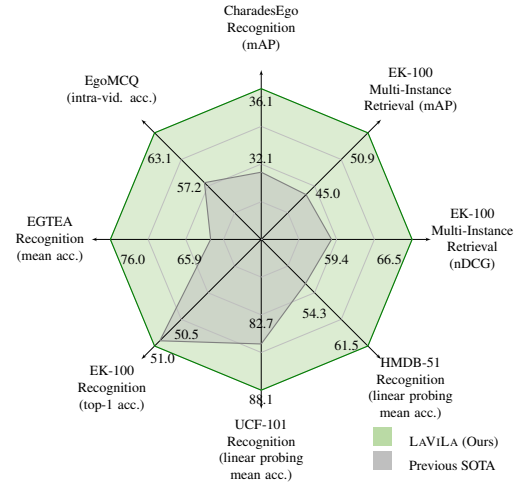

Learning visual representation using web-scale image-text data is %
a powerful tool for computer vision. Vision-language approaches~\cite{radford2021clip,yu2022coca,jia2021align} have pushed the state-of-the-art across a variety of tasks, including
zero-shot classification~\cite{radford2021clip}, novel object detection~\cite{zhou2022detecting}, %
and even image generation~\cite{ramesh2022hierarchical}.
Similar approaches %
for videos~\cite{bain2021frozen,nagrani2022learning,lin2022egovlp}, however, have been limited by the small size of paired video-text corpora compared to the billion-scale image-text datasets~\cite{radford2021clip,jia2021align,zhai2022scaling}---even though access to
raw video data has exploded in the past decade. %
In this work, we show it is possible to {\em automatically} generate text pairing for such videos by leveraging Large Language Models (LLMs), thus taking full advantage of the massive video data.
Learning video-language models with these automatically generated annotations leads to stronger representations,
and as~\cref{fig:teaser_sota} shows, sets a new state-of-the-art on six popular first and third-person video benchmarks.

\begin{figure}[t]
	\begin{center}
		\centering
		\includegraphics[width=\linewidth]{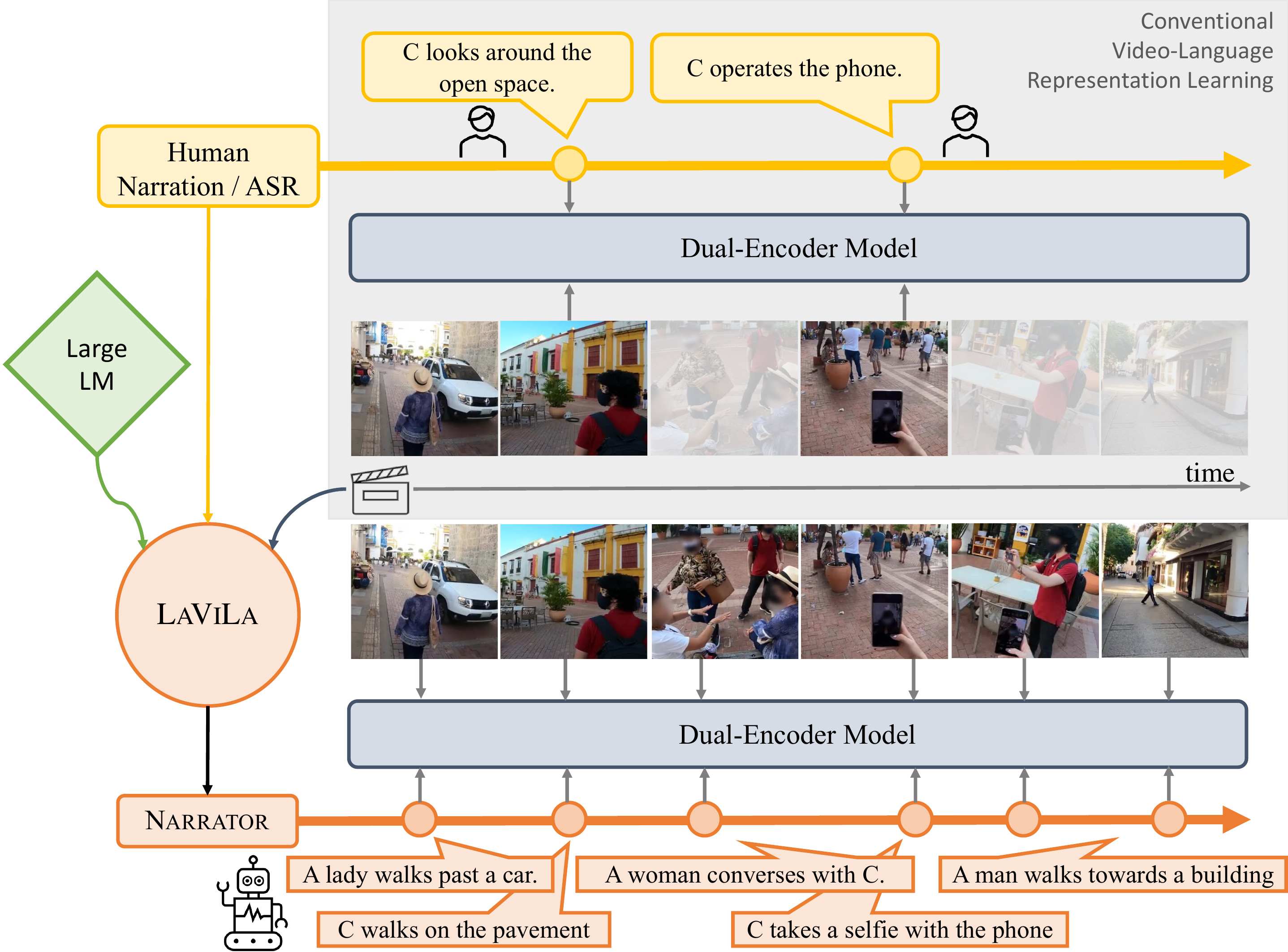}
		\caption{
			{\bf \ours} leverages Large Language Models (LLMs) to densely narrate long videos, and uses those narrations to train strong dual-encoder models. While prior work uses sparsely labeled text by humans, or weakly aligned text transcribed from speech, \ours is able to leverage dense, diverse, and well-aligned text generated by a LLM.
		}
		\label{fig:teaser}
	\end{center}
\end{figure}

Our method, called {\bf \ours}:
{\bf L}anguage-model {\bf a}ugmented {\bf Vi}deo-{\bf La}nguage pre-training,
leverages pre-trained LLMs, %
\eg GPT-2~\cite{radford2019gpt2},
which encode within their weights a treasure trove of factual knowledge and conversational ability.
As shown in~\cref{fig:teaser}, we repurpose these LLMs to be ``visually-conditioned narrators'', and finetune on all accessible paired video-text clips. Once trained,
we use the model to densely annotate thousands of hours of videos by generating rich textual descriptions.
This pseudo-supervision can thus pervade the entire video, in between and beyond the annotated snippets.
Paired with another LLM trained to rephrase existing narrations, \ours is able to create a much larger and more diverse set of text targets for video-text contrastive learning.
In addition to setting a new state-of-the-art as noted earlier, the stronger representation learned by \ours
even outperforms prior work using only half the groundtruth annotations (\cref{fig:percentage}).

\ours's strong performance can be attributed to a number of factors.
First, \ours can provide temporally dense supervision for long-form videos, where the associated captions are either too sparse, or the video-level ``Alt-Text'' (in the case of web videos) does not describe all the nuanced activities happening in it.
Second, the generated text is well-aligned with the visual input.
Although prior work has leveraged automatic speech transcription on How-To videos~\cite{miech2019howto100m} to automatically extract clips paired with text from the speech, %
such datasets %
have relatively poor alignment between the visual and textual content ($\le50\%$, \cf~\cite{miech2019howto100m,han2022tan}), limiting the quality of the learned representations.
Third, \ours can significantly expand annotations when only a little is available.
For instance, videos of mundane day-to-day activities, especially from an egocentric viewpoint, could be very useful for assistive and augmented reality applications.
Such videos, however, are rare on the internet, and hence do not readily exist with associated web text. Recent work~\cite{grauman2022ego4d} instead opted to manually capture and narrate such video data.
These narrations however required significant manual effort: 250K hours of annotator time spent in narrating 3.6K hours of video.
In contrast, \ours is able to automatically narrate each video multiple times and far more densely, and hence learns much stronger representations.

We extensively evaluate \ours across multiple video-text pre-training datasets and downstream tasks to validate its effectiveness.
Specifically, after being pre-trained on Ego4D, the largest egocentric video datasets with narrations, \ours can re-narrate the whole dataset 10$\times$ over. The resulting model learned on these expanded narrations sets a new state-of-the-art on
a wide range of downstream tasks across challenging datasets, including multi-instance video retrieval on Epic-Kitchens-100 ({\bf 5.9\%} absolute gain on mAP), multiple-choice question answering on Ego4D ({\bf 5.9\%} absolute gain on intra-video accuracy), and action recognition on EGTEA ({\bf 10.1\%} absolute gain on mean accuracy). It obtains gains both when evaluated for zero-shot transfer to the new dataset, as well as after fine-tuning on that dataset.
Similar gains are shown in third-person video data. When training \ours after densely re-narrating HowTo100M, we outperform prior work on downstream action classification on UCF-101 and HMDB-51.
In a case study of semi-supervised learning, we show that our model, which only ever sees 50\% of the human-labeled data, is capable of outperforming the baseline model trained with all the narrations.
Moreover, the gains progressively increase as we go to larger data regimes and larger backbones, suggesting the scalability of our method.

\section{Related Work}

\myparagraph{Vision-language representation learning}
maps visual and textual embeddings into a common space using metric-learning techniques~\cite{weston2010wsabie,frome2013devise}.
 Recently, different pretext tasks are proposed to learn a finer-grained association between visual and textual modality,~\eg masked language modeling (MLM)~\cite{lu2019vilbert,chen2020uniter,su2020vlbert} and captioning~\cite{desai2021virtex,yu2022coca}.
 Another line of research focuses on scaling up both models and pre-training data.
For instance, CLIP~\cite{radford2021clip} is pre-trained on 400M image-text pairs with a contrastive loss (InfoNCE~\cite{sohn2016improved,oord2018cpc}) while CoCa~\cite{yu2022coca} unifies contrastive and generative approaches with a single foundation model.
Similar trends are also witnessed in the video-text domain~\cite{sun2019videobert,zhu2020actbert,lei2021clipbert}.
However, collecting high-quality video-text data is more difficult than image-text.
Therefore, efforts are made to learn from uncurated videos with machine-generated audio transcripts via contrastive learning~\cite{miech2020milnce,xu2021videoclip,zellers2021merlot} or unsupervised alignment~\cite{han2022tan} while other works focus on either adapting well-performing image-text models to videos~\cite{lin2022frozen,ni2022expanding,xue2022clip,ju2022prompting}, or curriculum learning from a single frame to multiple frames~\cite{bain2021frozen}.
In contrast, our approach leverages language models to generate temporally dense textual supervision on long-form videos.

\myparagraph{Generative Visual Language Models (VLM)}
were first used for image/video captioning using recurrent networks~\cite{donahue2015lstm,vinyals2015show} and Transformer-based architectures~\cite{luo2020univl,seo2022end}.
More recently, generative VLMs have unified multiple vision tasks~\cite{cho2021vlt5,zhu2022uniperceiver} by training multi-modal Transformers on visual-text pairs~\cite{hu2022scaling,yuan2021florence}.
Meanwhile, generative VLMs also excel at multimodal tasks via zero-shot or few-shot prompting~\cite{tsimpoukelli2021multimodal,zeng2022socratic,alayrac2022flamingo} by leveraging multi-billion-parameter LLMs pre-trained on massive text corpus~\cite{radford2019gpt2,brown2020gpt3,hoffmann2022chinchilla}.
In our work, we demonstrate that generative VLMs can narrate long videos and the resulting video-text data benefits video-language representation learning.

\myparagraph{Large-scale multimodal video datasets} are crucial for video understanding tasks but are hard to collect.
Conventional video-text datasets~\cite{zhou2018youcook2,rohrbach2017lsmdc,caba2015activitynet} either have limited scenarios,~\eg cooking, or are not large enough to learn generic video representation.
Miech~\etal~\cite{miech2019howto100m} scrape over 100 million video-text pairs via automatic audio transcription from long-form How-To videos.
However, ASR introduces noticeable textual noise and visual-text unalignment~\cite{han2022tan}.
WebVid~\cite{bain2021frozen} contains 10 million short videos with textual descriptions.
But it is still several orders of magnitude smaller than the image counterparts~\cite{radford2021clip,schuhmann2022laion} and is harder to scale up since it is sourced from stock footage sites.
The recently released Ego4D~\cite{grauman2022ego4d} dataset offers 3,600 hours of egocentric videos in which written sentence narrations are manually annotated every few seconds but requires significant manual effort.
In contrast, our method shows a promising alternative by automatically narrating videos using supervision from LLM.

\myparagraph{Data augmentation techniques in NLP},
including word-level replacement based on synonyms~\cite{zhang2015character,wei2019eda} or nearest-neighbor retrieval~\cite{wang2015s,fadaee2017data}, improve text classification accuracy.
We refer readers to~\cite{feng2021survey} for a comprehensive survey.
In this paper, we show that sentence-level paraphrasing based on text-to-text models~\cite{raffel2020t5} is helpful for video-language pre-training.

\section{Preliminaries}
\label{sec:prelim}
A video $V$ is a stream of moving images $I$. %
The number of frames $|V|$ can be arbitrarily long
while video models typically operate on shorter clips, which are often in the range of a few seconds.
Therefore, we skim through a long-form video and represent it by a set of $N$ short clips,~\ie $\calX$.
Each clip $x_i$ is defined by a specific start and end frame $x_i = \{ I_{t_i}, \cdots, I_{e_i} \}$, where $0<\!t_i\!<\!e_i\!\leq\!|V|$, and is typically associated with some annotation $y_i$.
This annotation could be a class label or free-form textual description of the clip.
We denote a video by the set of annotated clips with their corresponding annotations,~\ie $(\calX,\calY) = \{ (x_1, y_1), \cdots,  (x_N, y_N) \}$.
Note that the annotated clips often cannot densely cover the entire video due to the annotation cost and visual redundancy, \ie $\bigcup_{i} [t_i, e_i] \subsetneq [0, |V|] $.

A typical video model $\mathcal{F} (\calX, \calY)$ learns from these clip-level annotations using a standard training objective such as a cross-entropy loss when the annotations are class labels with a fixed vocabulary. However, more recently, dual-encoder-based contrastive approaches like CLIP~\cite{radford2021clip,xu2021videoclip} have gained popularity.
They work with free-form textual annotations which are tokenized~\cite{sennrich2016bpe} into sequences of discrete symbols,~\ie $ y=(s_1, s_2, \cdots, s_{L})\in \{ 1, 0 \}^{ |\mathbb{S}| \times L } $.
The model consists of a visual encoder $f_\mathrm{v}: \mathbb{R}^{T\times3\times H\times W} \mapsto \mathbb{R}^{D_\mathrm{v}}$ plus a projection head $h_\mathrm{v}: \mathbb{R}^{D_\mathrm{v}} \mapsto \mathbb{R}^{d} $ and a textual encoder $f_\mathrm{t}: \{ 1, 0 \}^{ |\mathbb{S}| \times L } \mapsto \mathbb{R}^{D_\mathrm{t}} $ plus a projection head $ h_\mathrm{t}: \mathbb{R}^{D_\mathrm{t}} \mapsto \mathbb{R}^{d} $ in parallel to obtain the global visual and textual embedding respectively:
{\small
\begin{align*}
	\mathbf{v} = h_\mathrm{v}(f_\mathrm{v}(x)), \quad
	\mathbf{u} = h_\mathrm{t}(f_\mathrm{t}(y)).
\end{align*}
}
A contrastive loss, such as InfoNCE~\cite{oord2018cpc}, learns global embeddings that associate corresponding video and text embeddings within a batch of samples $\mathcal{B}$,
{\small
\begin{align}
	\frac{1}{ |\mathcal{B}| } \sum_{(x,y)\in \mathcal{B}} \left( \mathrm{InfoNCE}(\mathbf{v}, \mathbf{u}) + \mathrm{InfoNCE}(\mathbf{u}, \mathbf{v}) \right).
\end{align}
}

\begin{figure}
	\centering
	\includegraphics[width=\linewidth]{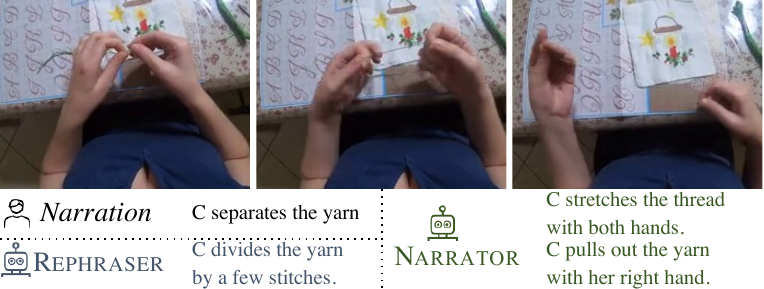}
	\includegraphics[width=\linewidth]{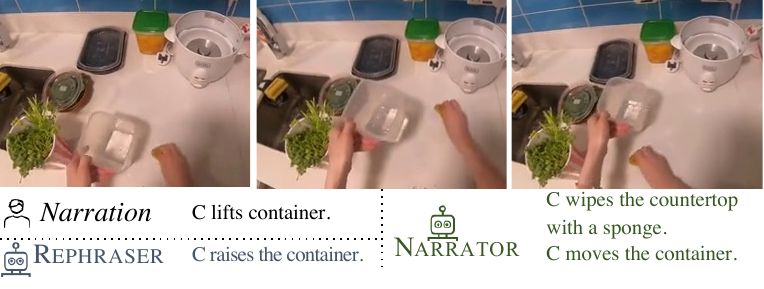}
	\caption{
		\textbf{Generated samples by our \narrator and \rephraser}.
		\narrator generates new descriptions of the action taking place, potentially focusing on other objects being interacted with.
		\rephraser not only changes the word order of the human narration but also diversifies it by using related verbs or nouns.
	}
	\label{fig:qual}
\end{figure}

\section{\ours}

In \ours, we leverage large language models (LLMs) as supervision to train the dual-encoder model, where the LLMs serve as vision-conditioned narrators and automatically generate textual descriptions from video clips.
In particular, we exploit supervision from two LLMs:
(1) {\bf \narrator} (\cref{sec:method:pseudo_caption}) is a {\em visually-conditioned} LLM that pseudo-labels existing and new clips with narrations, generating new annotations $(\calX', \calY')$.
(2) {\bf \rephraser} (\cref{sec:method:paraphrase}) is a standard LLM that paraphrases narrations in existing clips, augmenting those annotations to $(\calX, \calY'')$.
As illustrated in~\cref{fig:qual}, \narrator generates new descriptions of the action taking place, potentially focusing on other objects being interacted with.
\rephraser serves to augment the text input, \eg, changes word order of the human narration and additionally replaces common verbs or nouns, making annotations more diverse.
Finally, we train the {\bf dual-encoders} (\cref{sec:method:lavie}) on all these annotations combined,
\ie $(\calX, \calY) \cup (\calX', \calY') \cup (\calX, \calY'')$.

\begin{figure*}[t]
	\vspace{-10pt}
	\centering
	\includegraphics[width=0.95\linewidth]{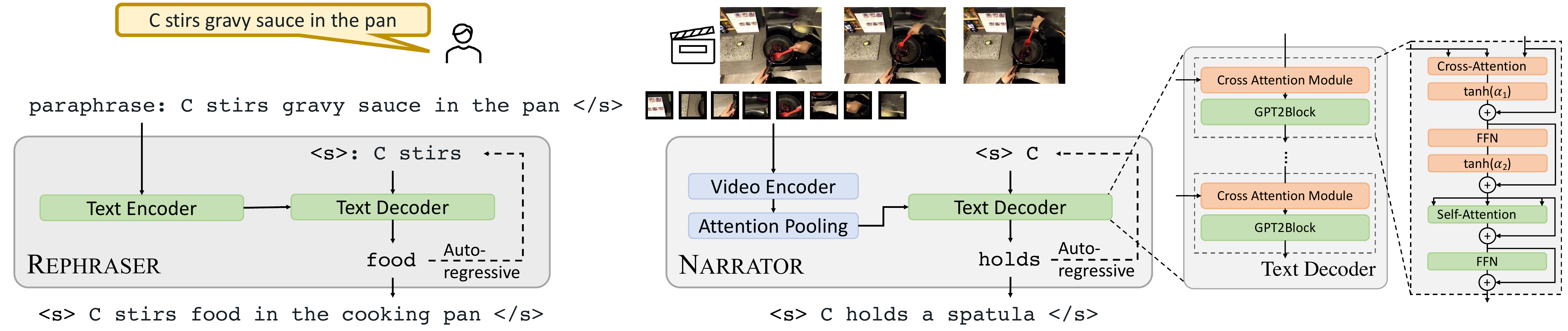}
	\caption{{\bf Language supervision from \rephraser and \narrator}.
	\rephraser {\em (left)} takes the narration as input, passes it through a text encoder and uses a text decoder to autoregressively generate the rephrased output.
	\narrator {\em (right)} takes video frames as input and obtains the visual embeddings through a video encoder followed by attentional pooling. Equipped with a few additional cross-attention modules, the text decoder autoregressively generates new narrations for those new frames.}
	\label{fig:method}
\end{figure*}

\subsection{\bf \narrator}
\label{sec:method:pseudo_caption}

Traditional LLMs, such as GPT-2~\cite{radford2019gpt2}, are trained to generate a sequence of text tokens $(s_1\cdots\!s_{L})$ from scratch by modeling the probability of the next token given all tokens seen so far: $p(s_l | s_{<l})$.
\narrator repurposes existing LLMs to be conditioned on the visual input and is trained on the original annotations $(\calX, \calY)$. The resulting model produces dense new annotations $(\calX', \calY')$ on the full video.
Following the formulation of factorized probabilities in language models~\cite{bengio2020neural}, we model the visually conditioned text likelihood as follows:
{\small
\begin{align}
	\label{eq:vlm}
	p_\mathrm{\narrator}(y' \vert x') = \prod_{\ell=1}^L p(s'_\ell | s'_{<\ell} , x').
\end{align}
}

\myparagraph{Architecture.}
We design \narrator to closely follow the architecture of standard LLMs, with only a few additional cross-attention modules added to provide visual conditioning, as illustrated in \cref{fig:method} ({\em right}).
This enables \narrator to be initialized from pre-trained weights, which is crucial for our task as the data we use to train \narrator (narrations associated with video clips) are far smaller in scale
compared to the large text corpus typically used to train LLMs. Moreover, video narrations are less diverse and noisier because they are either collected by only a few annotators or automatically transcribed from speech.
Similar ``frozen-LM'' approaches have shown effectiveness in
multimodal few-shot adaptation in recent work~\cite{tsimpoukelli2021multimodal,alayrac2022flamingo}.

Specifically, we take a frozen pre-trained LLM and add a cross-attention module before each Transformer decoder layer so that the text input can attend to visual information.
The cross-attended output then sums with the input text feature via residual connection~\cite{he2016resnet} and goes to the Transformer decoder layer.
Each cross-attention module comprises a cross-attention layer,
which takes textual tokens as queries and visual embedding as keys and values, followed by a feed-forward network (FFN).
Layer Normalization~\cite{ba2016layernorm} is applied at the beginning of both cross-attention and FFN.
We add {\tt tanh}-gating~\cite{hochreiter1997lstm}, with an initial value of zero, such that the output of the new model is the same as that from the original language model at the beginning.

While features from any video model are applicable for conditioning, for convenience we adopt the video encoder from $\mathcal{F}$ in~\cref{sec:prelim}, trained
contrastively %
on the ground-truth data $(\calX, \calY)$. We use features before global pooling to allow the LLM to leverage fine-grained spatial-temporal information.

\myparagraph{Training.}
We train \narrator on all of, or a subset of, the ground-truth annotations $(\calX, \calY)$.
For each pair $(x, y)$, the captioning loss is the sum of the negative log-likelihood of the correct word at each step,
{\small
\begin{align}
	\mathcal{L}_\mathrm{\narrator} (x, y) = -\sum_{\ell=1}^L \log p(s_\ell | s_{<\ell}, x).
\end{align}
}

\myparagraph{Inference.}
At inference time, we query \narrator by feeding visual input $x$ plus a special start-of-sentence token \texttt{<s>}. We sample from the distribution recursively,~\ie $\tilde{s}_\ell \sim p(s| [\texttt{<s>}, \cdots, \tilde{s}_{\ell-1}],x)$ until an end-of-sentence token \texttt{</s>} is reached.
Following~\cite{holtzman2020curious},
at each step we sample from a subset of tokens that contain the vast majority of the probability mass, which is known as nucleus sampling.

The effect of nucleus sampling is two-fold.
On the one hand, it generates more diverse, open-ended, and human-like text than maximum-likelihood-based methods such as beam search and its variants~\cite{vijayakumar2016diverse}.
On the other hand, the generated text may contain irrelevant or noisy information due to sampling without post-processing based on sentence-level likelihood.
To address this, we repeat the sampling process for $K$ times on the same visual input.
We later demonstrate that the contrastive pre-training objective is robust to the noise caused by sampling, and the final performance benefits from a more diverse set of narrations. %

To sample video clips for captioning, we start by simply re-captioning the existing clips labeled in the dataset $\calX$, resulting in expanded annotations. Furthermore, long-form videos are typically sparsely narrated, meaning that the temporal union of all labeled clips cannot cover the entire video.
Hence, we use \narrator to annotate the remainder of the video to obtain additional annotations by pseudo-captioning.
With a simple assumption that video is a stationary process, we uniformly sample clips from the unlabeled intervals.
The clip duration is equal to the average of all ground-truth clips,~\ie $\Delta = \frac{1}{N}\sum_{i=1}^N(e_i - t_i) $ while the sampling stride is computed likewise.
Finally, by combining both re-captioned and pseudo-captioned narrations, %
we refer to the final set of annotations generated by \narrator as $(\calX', \calY')$.

\myparagraph{Post-processing.}
Exhaustive pseudo-captioning may contain some uninformative visual clips and generate text that is not useful.
Thus, we add a filtering process to eliminate low-quality clips and their associated descriptions.
We use the baseline dual-encoder model $\mathcal{F}$, which is trained on the ground-truth paired clips, to compute the visual and textual embedding of pseudo-labeled pairs and filter based on the similarity score, \ie $\mathrm{Filter}(f_\mathrm{v}(x'_j)^\top \cdot f_\mathrm{t}(y'_j))$, where $\mathrm{Filter}(\cdot)$ can be either top-$k$ of all generated text or a threshold filtering.
In the experiments, we use a threshold of $0.5$.

\subsection{\bf \rephraser}
\label{sec:method:paraphrase}
The data generated by \narrator is several times larger than the ground-truth pairs.
To ensure that we do not overfit the pseudo-labeled data, we increase the number of ground-truth narrations by paraphrasing.
In particular, we use a text-to-text LLM which models conditional text likelihood:
{\small
\begin{align*}
	p_\mathrm{\rephraser}(y''\vert y) = \prod_{\ell=1}^L p(s''_\ell \vert s''_{<\ell}, y).
\end{align*}
}
The text-to-text model is implemented by an encoder-decoder architecture,~\eg T5~\cite{raffel2020t5}, to auto-regressively generate a new sentence given the original one.
We observe that \rephraser is able to do basic manipulations such as replacing synonyms or changing word order, which serves as an efficient way of automatic data augmentation. The resulting annotations are referred to as $(\calX, \calY'')$.

\subsection{Training the Dual-Encoders}
\label{sec:method:lavie}
We train the dual-encoders as described in~\Cref{alg:lavie} in \cref{sec:appdx:results}.
In each iteration, we first sample a batch $\mathcal{B}$ of video clips. It comprises a subset of clips $\mathcal{B}_l$ with labeled timestamps as well as narrations, and a subset $\mathcal{B}_u$ whose clips are randomly sampled from videos without narrations.
For clip $x_i\in\mathcal{B}_u$, we obtain the pseudo-caption $y'_i$ by querying the \narrator $y'_i \sim p_\mathrm{\narrator}(y'|x)$, resulting in a set of clips with LLM-generated narrations $\mathcal{\widetilde{B}}_u$.
For clip $(x_i, y_i)\in\mathcal{B}_l$, the text supervision is obtained from either the \rephraser or the \narrator, with a probability of $0.5$.
We denote the resulting set of pairs to be $\mathcal{\widetilde{B}}_l$ similarly.
Following CLIP~\cite{radford2021clip}, we use the symmetric cross-entropy loss over the similarity scores of samples in the batch $\mathcal{\widetilde{B}}_l \cup \mathcal{\widetilde{B}}_u$.

In practice, we run \rephraser and \narrator in advance and cache the resulting video-narration pairs so that there is no computational overhead during pre-training.
Therefore, training a dual-encoder in \ours is as fast as training a standard dual-encoder contrastive model.

\begin{table}
	\tablestyle{2pt}{1.05}
	\resizebox{\linewidth}{!}{
	\begin{tabular}{y{60}|x{24}|x{16}|x{72}|x{36}}
	Datasets & Task & Ego? & Metrics & Eval. Prot. \\
	\thickhline
	\ek~\cite{damen2022epickitchens100} &MIR & \cmark & mAP, nDCG & ZS, FT \\
	\ek~\cite{damen2022epickitchens100} & CLS & \cmark & top-1 action acc. & FT \\
	Ego4D~\cite{grauman2022ego4d} & MCQ & \cmark & {\scriptsize inter-/intra-video acc.} & ZS \\
	Ego4D~\cite{grauman2022ego4d} & NLQ & \cmark & Recall@N & FT \\
	EGTEA~\cite{li2018egtea} & CLS & \cmark & top-1, mean acc. & ZS, FT \\
	\charadesego~\cite{sigurdsson2018charadesego} & CLS & \cmark & video-level mAP & ZS, FT \\
	UCF-101~\cite{soomro2012ucf101} & CLS & \xmark & mean acc. & LP \\
	HMDB-51\tablefootnote{HMDB data is licensed under the CC BY 4.0 license and the data is available at 
	\url{https://serre-lab.clps.brown.edu/resource/hmdb-a-large-human-motion-database/}}~\cite{kuehne2011hmdb} & CLS & \xmark & mean acc. & LP \\
	\hline
	\end{tabular}}
\caption{\textbf{Downstream datasets} and metrics used for evaluation.
We evaluate \ours on a wide range of tasks including Multi-Instance Retrieval (MIR), Multiple-Choice Question (MCQ), Natural Language Query (NLQ), and Action Recognition (CLS).
The evaluation protocols include zero-shot (ZS), fine-tuning (FT), and linear-probing (LP).
Please refer to \cref{sec:appdx:dataset} for more details.
}
\label{tab:datasets}
\end{table}

\section{Experiments}

\myparagraph{Dual-Encoder Architecture.}
The video-language model follows a dual-encoder architecture as CLIP~\cite{radford2021clip}.
The Visual encoder is a TimeSformer (TSF)~\cite{bertasius2021timesformer}, whose spatial attention modules are initialized from a ViT~\cite{dosovitskiy2020vit}
which is contrastively pre-trained on large-scale paired image-text data as in CLIP~\cite{radford2021clip}.
We sample 4 frames per clip during pre-training and 16 when finetuning on downstream tasks.
The text encoder is a 12-layer Transformer~\cite{vaswani2017attention,radford2019gpt2}.
We use BPE tokenizer~\cite{sennrich2016bpe} to pre-process the full sentence corresponding to the video clip and keep at most 77 tokens.

\myparagraph{\narrator}'s
architecture is a visually conditioned auto-regressive Language Model.
The visual encoder is by default TimeSformer-L while the text decoder is a GPT-2 XL.
During inference, we use nucleus sampling~\cite{holtzman2020curious} with $p=0.95$ and return $K=10$ candidate outputs.

\myparagraph{\rephraser.}
We use an open-source paraphraser~\cite{golla2021rephraser} %
 based on T5-large~\cite{raffel2020t5}.
It is pre-trained on C4~\cite{raffel2020t5} and then finetuned on a cleaned subset of ParaNMT~\cite{wieting2018paranmt}.
During inference, we use Diverse Beam Search~\cite{vijayakumar2016diverse} with group number the same as beam number ($G=B=20$) and set the diversity penalty to be 0.7.
We keep 3 candidates per sentence, remove punctuations, and do basic de-duplication.

\myparagraph{Pre-training dataset.}
We train on the video-narration pairs from Ego4D~\cite{ego4d,grauman2022ego4d}, the largest egocentric video dataset to date.
We exclude videos that appear in the validation and test sets of the Ego4D benchmark and determine each clip's interval using the same pairing strategy in~\cite{lin2022egovlp}.
This results in around 4M video-text pairs with an average clip length of 1 second.
We also experiment with third-person videos by pre-training on HowTo100M~\cite{miech2019howto100m} in \cref{sec:expt:third_person}.

\begin{table}
	\tablestyle{2pt}{1.05}
	\begin{tabular}{y{48}H|x{36}|x{20}x{20}x{20}|x{20}x{20}x{20}}
		\multirow{2}{*}{Method} & \# frames & \multirow{2}{*}{Backbone} &  \multicolumn{3}{c}{mAP} & \multicolumn{3}{|c}{nDCG} \\
		& & & V$\rightarrow$T & T$\rightarrow$V & Avg. & V$\rightarrow$T & T$\rightarrow$V & Avg. \\
		\thickhline
		\multicolumn{9}{l}{(\textsc{Zero-shot})} \\
		\hline
		EgoVLP~\cite{lin2022egovlp} & 4 & TSF-B & 19.4 & 13.9 & 16.6 & 24.1 & 22.0 & 23.1  \\
		EgoVLP$^*$~\cite{lin2022egovlp} & 16 & TSF-B & \underline{26.0} & \underline{20.6} & \underline{23.3} & \underline{28.8} & \underline{27.0} & \underline{27.9}  \\
		\arrayrulecolor{LightGrayForTableRule}
		\hline
		\arrayrulecolor{Black}
		\ours & 4 & TSF-B & {\bf 35.1} & {\bf 26.6} & {\bf 30.9} & {\bf 33.7} & {\bf 30.4} & {\bf 32.0} \\
		\rowcolor{Gray} \ours & 4 & TSF-L  & {\bf 40.0} & {\bf 32.2} & {\bf 36.1} & {\bf 36.1} & {\bf 33.2} & {\bf 34.6} \\
		\hline
		\multicolumn{9}{l}{(\textsc{Finetuned})} \\
		\hline
		MME~\cite{wray2019jpose} & 25 & TBN & 43.0 & 34.0 & 38.5 & 50.1 & 46.9 & 48.5 \\
		JPoSE~\cite{wray2019jpose} & 25 & TBN & \underline{49.9} & 38.1 & 44.0 & 55.5 & 51.6 & 53.5 \\
		EgoVLP~\cite{lin2022egovlp} & 16 & TSF-B & \underline{49.9} & \underline{40.5} & \underline{45.0} & \underline{60.9} & \underline{57.9} & \underline{59.4}  \\
		\arrayrulecolor{LightGrayForTableRule}
		\hline
		\arrayrulecolor{Black}
		\ours & 16 & TSF-B & {\bf 55.2} & {\bf 45.7} & {\bf 50.5} & {\bf 66.5} & {\bf 63.4} & {\bf 65.0} \\
		\rowcolor{Gray} \ours & 16 & TSF-L & {\bf 54.7} & {\bf 47.1} & {\bf 50.9} & {\bf 68.1} & {\bf 64.9} & {\bf 66.5} \\
		\hline
	\end{tabular}
	\caption{\textbf{\ekmir}.
		\ours outperforms prior work across all settings, metrics and directions of retrieval, with larger gains when switching to a larger model.
		Specifically, our best model achieves over $10\%$ absolute gain in the zero-shot setting and $5.9\sim 7.1\%$ gain in the finetuned setting.
		EgoVLP$^*$ refers to our improved version of~\cite{lin2022egovlp}, details of which are given %
        in \cref{sec:appdx:ablations}.
	}
	\label{tab:sota_ek100}
\end{table}

\begin{table}
	\tablestyle{2pt}{1.05}
	\begin{tabular}{y{48}|x{36}x{36}|x{20}x{20}x{20}x{20}}
		\multirow{3}{*}{Method} & \multicolumn{2}{c|}{\egomcq} & \multicolumn{4}{c}{\egonlq} \\
		& \multicolumn{2}{c|}{Accuracy (\%)} & \multicolumn{2}{c}{mIOU@0.3} & \multicolumn{2}{c}{mIOU@0.5} \\
		& Inter-video & Intra-video &  R@1 & R@5 & R@1 & R@5 \\
		\thickhline
		SlowFast~\cite{grauman2022ego4d} & - & - & 5.45 & 10.74 & 3.12 & 6.63 \\
		EgoVLP~\cite{lin2022egovlp} & \underline{90.6} & \underline{57.2} & {\bf 10.84} & \underline{18.84} & {\bf 6.81} & \underline{13.45}  \\
		\arrayrulecolor{LightGrayForTableRule}
		\hline
		\arrayrulecolor{Black}
		\ours (B) & {\bf 93.8} & {\bf 59.9} & \underline{10.53} & {\bf 19.13} & \underline{6.69} & {\bf 13.68}  \\
		\rowcolor{Gray} \ours (L) & {\bf 94.5} & {\bf 63.1} & {\bf 12.05} & {\bf 22.38} & {\bf 7.43} & {\bf 15.44} \\
		\arrayrulecolor{Black}
		\hline
	\end{tabular}
	\caption{\textbf{Ego4D \egomcq and \egonlq.} \ours outperforms prior work on both Multiple-Choice Questions and Natural Language Questions on Ego4D, with nearly 6\% absolute gain on the challenging intra-video MCQ task that requires reasoning over multiple clips from the same video to answer a question.}
	\label{tab:sota_ego4d_mcq}
\end{table}

\myparagraph{Evaluation protocols.} We evaluate the learned video-text encoders using three evaluation protocols. (1) {\em Zero-Shot (ZS)}, meaning that we apply the pre-trained video-text encoders directly on the downstream validation dataset to perform video$\leftrightarrow$text retrieval tasks, without any tuning on the downstream dataset.
Zero-shot classification is performed similarly,
where we compute the similarity score between the video clip and the textual description of all possible categories.
(2) {\em Finetuned (FT)}, where we take the pre-trained video-text model and perform end-to-end fine-tuning on the training split of the target downstream dataset.
(3) {\em Linear-Probe (LP)}, where we compute the video features from a frozen encoder and train a linear SVM on top of the training split of the downstream dataset.

\myparagraph{Downstream benchmarks.}
We %
use multiple benchmarks across four first-person (egocentric) and two third-person datasets, as enumerated in~\cref{tab:datasets}. We summarize them here and refer the reader to~\cref{sec:appdx:dataset} for details on datasets and metrics.
(1) Two tasks on Epic-Kitchens-100: Multi-Instance Retrieval ({\bf \ekmir}) and Action Recognition ({\bf \ekcls})~\cite{damen2022epickitchens100}. \ek is a very popular and challenging egocentric video recognition benchmark. The MIR task requires retrieving the text given videos (V$\rightarrow$T) and videos given text (T$\rightarrow$V).
The CLS task requires classifying each video into one of 97 verbs and 300 nouns each, resulting in a combination of 3,806 action categories.
(2) Two downstream tasks of Ego4D: Multiple-Choice Questions ({\bf \egomcq}) and Natural Language Query ({\bf \egonlq}).
\egomcq requires selecting the correct textual description from five choices given a query video clip while \egonlq asks the model to output the relevant temporal intervals of video given a text query.
We select these two benchmarks because they require reasoning about both visual and textual information.
(3) Action Recognition on {\bf EGTEA}~\cite{li2018egtea}. It requires classifying into 106 classes of fine-grained cooking activities.
(4) Action Recognition on {\bf CharadesEgo}~\cite{sigurdsson2018charadesego}.
It requires classification into 157 classes of daily indoor activities.
Note that CharadesEgo is very different from \ek, Ego4D and EGTEA since its videos are captured by head-mounted phone cameras in a crowd-sourcing way.

\begin{table}
	\tablestyle{2pt}{1.05}
\begin{tabular}{y{54}|x{40}|x{48}|x{36}x{36}}
	Method &  Backbone & Pretrain & Top-1 Acc. & Mean Acc. \\
	\thickhline
	Li~\etal~\cite{li2018egtea} & I3D & K400 & - & 53.30 \\
	LSTA~\cite{sudhakaran2019lsta} & ConvLSTM & IN-1k & 61.86 & 53.00 \\
	IPL~\cite{wang2021ipl} & I3D & K400 & - & 60.15 \\
	MTCN~\cite{kazakos2021little} & {\tiny SlowFast (V+A+T)} & {\tiny K400+VGG-Sound} & \underline{73.59} & \underline{65.87} \\
	\arrayrulecolor{LightGrayForTableRule}
	\hline
	\arrayrulecolor{Black}
	Visual only & TSF-B & IN-21k+K400 & 65.58 & 59.32 \\
 	\ours & TSF-B & WIT+Ego4D & {\bf 77.45} & {\bf 70.12} \\
 	\rowcolor{Gray} \ours & TSF-L & WIT+Ego4D & {\bf 81.75} & {\bf 76.00} \\
 	\hline
\end{tabular}
\caption{\textbf{EGTEA Classification}. \ours obtains significant gains on this task, outperforming prior work with over $10\%$ mean accuracy. Since the backbones used are not all comparable, we also report a comparable baseline with TSF-B (``Visual only'').
}
\label{tab:sota_egtea}
\end{table}

In all tables, we bold and underline the best and second-best performing methods with comparable backbones architectures. We \hlGray{highlight} the overall best performing method, which typically uses a larger backbone, if applicable.

\subsection{Main Results}
\label{sec:expt:main}

\myparagraph{\ek.}
We compare \ours with prior works on \ekmir in \cref{tab:sota_ek100}.
In the zero-shot setup, \ours remarkably surpasses an improved version of EgoVLP~\cite{lin2022egovlp} under similar model complexity: we use TSF-Base+GPT-2 as the dual-encoder architecture while EgoVLP uses TSF-Base+Distil-BERT.
With a stronger video encoder,~\ie TSF-Large, the performance improves further.
In the fine-tuned setting, \ours significantly outperforms all previous supervised approaches including MME, JPOSE~\cite{wray2019jpose} and EgoVLP~\cite{lin2022egovlp}. We also compare \ours on \ekcls in~\cref{sec:appdx:results}, and establish a new state-of-the-art.

\begin{table}
	\tablestyle{2pt}{1.05}
	\begin{tabular}{y{80}|x{60}|x{36}|x{36}}
		Method &  Backbone  & mAP (ZS)  & mAP (FT)  \\
		\thickhline
		ActorObserverNet~\cite{sigurdsson2018actor} & ResNet-152 & - & 20.0 \\
		SSDA~\cite{choi2020ssad} & I3D & - & 25.8 \\
		Ego-Exo~\cite{li2021egoexo} & SlowFast-R101 & - & 30.1 \\
		EgoVLP~\cite{lin2022egovlp} & TSF-B & \underline{25.0} & \underline{32.1} \\
		\arrayrulecolor{LightGrayForTableRule}
		\hline
		\arrayrulecolor{Black}
		\ours & TSF-B & {\bf 26.8} & {\bf 33.7} \\
		\rowcolor{Gray} \ours & TSF-L & {\bf 28.9} & {\bf 36.1} \\
		\hline
	\end{tabular}
	\caption{\textbf{CharadesEgo Action Recognition}. \ours sets new state-of-the-art in both zero-shot (ZS) and finetuned (FT) settings. Note that CharadesEgo videos are visually different compared to Ego4D videos, on which \ours is pretrained.}
	\label{tab:sota_charadesego}
\end{table}

\begin{figure*}[!b]
	\centering
	\begin{subfigure}[b]{\linewidth}
		\centering
		\begin{tikzpicture}
			\begin{customlegend}
				[
				legend columns=3, legend style={column sep=2ex},
				legend entries={\ours,Baseline,SOTA~\cite{lin2022egovlp}
				}]
				\addlegendimage{mark=triangle,style={thick},NiceBlue}
				\addlegendimage{mark=o,style={thick},gray}
				\addlegendimage{mark=asterisk,color_ao,mark options={scale=2,thick},only marks}
			\end{customlegend}
		\end{tikzpicture}
	\end{subfigure}

	\begin{subfigure}[b]{0.246\linewidth}
		\resizebox{\textwidth}{!}{
			\begin{tikzpicture}
	\begin{axis} [
		axis x line*=bottom,
		axis y line*=left,
		legend pos=north east,
		ymin=19, ymax=31,
		xmin=0, xmax=100,
		xticklabel={\pgfmathparse{\tick}\pgfmathprintnumber{\pgfmathresult}\%},
		xtick={0,20,50,100},
		ytick={21,23,25,27,29},
		width=\linewidth,
		legend style={cells={align=left}},
		label style={font=\footnotesize},
		tick label style={font=\footnotesize},
		legend style={at={(0.35,0.2)},anchor=west},
		]

		\addplot[mark=o,style={thick},gray] plot coordinates {
			(10, 21.8)
			(20, 23.5)
			(50, 25.4)
			(100, 26.0)
		};
		\addplot[mark=triangle,style={thick},NiceBlue] plot coordinates {
			(10, 22.7)
			(20, 24.6)
			(50, 25.8)
			(100, 28.2)
		};
		\addplot[dashed,gray] plot coordinates {
			(0, 26.0)
			(100, 26.0)
		};
		\addplot[mark=asterisk,color_ao,mark options={scale=2,thick},only marks] plot coordinates {
			(100, 22.3)
		};
		\label{pgf:ek100_map}
	\end{axis}
\end{tikzpicture}%
		}
		\caption{\ekmir mAP.}
	\end{subfigure}
	\begin{subfigure}[b]{0.246\linewidth}
		\resizebox{\textwidth}{!}{
			\begin{tikzpicture}
	\begin{axis} [
		axis x line*=bottom,
		axis y line*=left,
		legend pos=north east,
		ymin=26, ymax=32,
		xmin=0, xmax=100,
		xticklabel={\pgfmathparse{\tick}\pgfmathprintnumber{\pgfmathresult}\%},
		xtick={0,20,50,100},
		ytick={27,28,29,30,31},
		width=\linewidth,
		legend style={cells={align=left}},
		label style={font=\footnotesize},
		tick label style={font=\footnotesize},
		legend style={at={(0.35,0.2)},anchor=west},
		]

		\addplot[mark=o,style={thick},gray] plot coordinates {
			(10, 26.9)
			(20, 27.9)
			(50, 28.5)
			(100, 29.0)
		};
		\addplot[mark=triangle,style={thick},NiceBlue] plot coordinates {
			(10, 27.7)
			(20, 28.6)
			(50, 29.0)
			(100, 30.2)
		};
		\addplot[dashed,gray] plot coordinates {
			(0, 29.0)
			(100, 29.0)
		};
		\addplot[mark=asterisk,color_ao,mark options={scale=2,thick},only marks] plot coordinates {
			(100, 27.4)
		};
		\label{pgf:ek100_ndcg}
	\end{axis}
\end{tikzpicture}%
		}
		\caption{\ekmir nDCG.}
	\end{subfigure}
	\begin{subfigure}[b]{0.246\linewidth}
		\resizebox{\textwidth}{!}{
			\begin{tikzpicture}
	\begin{axis} [
		axis x line*=bottom,
		axis y line*=left,
		legend pos=north east,
		ymin=20, ymax=32,
		xmin=0, xmax=100,
		xticklabel={\pgfmathparse{\tick}\pgfmathprintnumber{\pgfmathresult}\%},
		ytick={22,24,26,28,30},
		xtick={0,20,50,100},
		width=\linewidth,
		legend style={cells={align=left}},
		label style={font=\footnotesize},
		tick label style={font=\footnotesize},
		legend style={at={(0.35,0.2)},anchor=west},
		title style={font=\Large},
		]
		\addplot[mark=o,style={thick},gray] plot coordinates {
			(10, 22.5)
			(20, 24.1)
			(50, 26.1)
			(100, 27.3)
		};
		\addplot[mark=triangle,style={thick},NiceBlue] plot coordinates {
			(10, 22.4)
			(20, 24.8)
			(50, 27.6)
			(100, 29.2)
		};
		\addplot[dashed,gray] plot coordinates {
			(0, 27.3)
			(100, 27.3)
		};
		\label{pgf:egtea_mean}
	\end{axis}
\end{tikzpicture}%
		}
		\caption{EGTEA mean accuracy.}
	\end{subfigure}
	\begin{subfigure}[b]{0.246\linewidth}
		\resizebox{\textwidth}{!}{
			\begin{tikzpicture}
	\begin{axis} [
		axis x line*=bottom,
		axis y line*=left,
		legend pos=north east,
		ymin=48, ymax=60,
		xmin=0, xmax=100,
		xticklabel={\pgfmathparse{\tick}\pgfmathprintnumber{\pgfmathresult}\%},
		xtick={0,20,50,100},
		ytick={50,52,54,56,58},
		width=\linewidth,
		legend style={cells={align=left}},
		label style={font=\footnotesize},
		tick label style={font=\footnotesize},
		legend style={at={(0.35,0.2)},anchor=west},
		title style={font=\Large},
		]
		\addplot[mark=o,style={thick},gray] plot coordinates {
			(10, 49.93)
			(20, 51.93)
			(50, 54.63)
			(100, 54.28)
		};
		\addplot[mark=triangle,style={thick},NiceBlue] plot coordinates {
			(10, 50.37)
			(20, 52.27)
			(50, 55.21)
			(100, 58.31)
		};
		\addplot[dashed,gray] plot coordinates {
			(0, 54.28)
			(100, 54.28)
		};
		\addplot[mark=asterisk,color_ao,mark options={scale=2,thick},only marks] plot coordinates {
			(100, 57.2)
		};
		\label{pgf:ego4d_mcq}
	\end{axis}
\end{tikzpicture}%
		}
		\caption{\egomcq Intra-video accuracy.}
	\end{subfigure}
	\caption{\textbf{\ours is effective in a semi-supervised setting where only a limited amout of narrations are given}.
	Comparing zero-shot performance of pre-training, \ours consistently outperforms the groundtruth-only baseline when 10, 20, 50, 100\% data is used.
	We also achieve comparable result with state-of-the-art with only 50\% of the annotated data.
	}
	\label{fig:percentage}
\end{figure*}
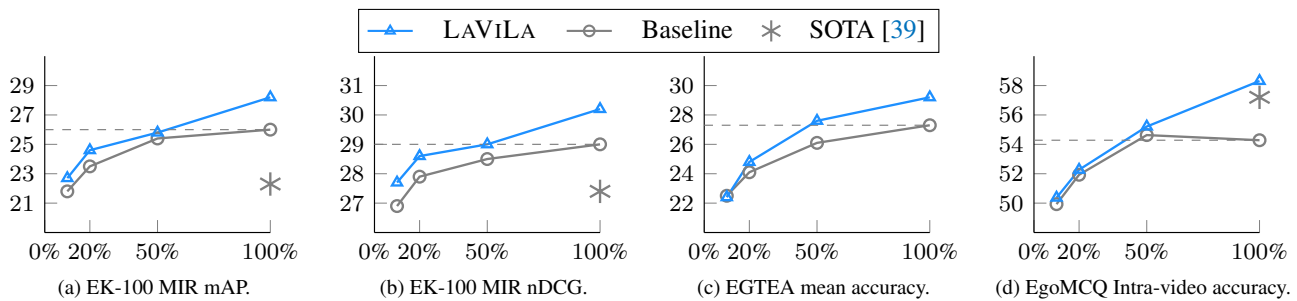

\myparagraph{Ego4D.}
We evaluate the pre-trained \ours model on \egomcq and \egonlq tasks and compare the results in \cref{tab:sota_ego4d_mcq}.
On \egomcq, our method achieves 93.8\% inter-video accuracy and 59.9\% intra-video accuracy, outperforming EgoVLP by a noticeable margin.
Note that EgoVLP's performance reported in~\cref{tab:sota_ego4d_mcq} is obtained by using EgoNCE loss~\cite{lin2022egovlp}, a variant of InfoNCE specialized for Ego4D while ours uses a standard InfoNCE loss.
EgoVLP with InfoNCE has lower performance (89.4\% inter-video and 51.5\% intra-video accuracy).
On \egonlq, \ours achieves comparable results with EgoVLP with similar model complexity.

\myparagraph{EGTEA.}
We evaluate the learned video representation by finetuning the video encoder for action classification in \cref{tab:sota_egtea} on another popular egocentric dataset, EGTEA~\cite{li2018egtea}.
Our method surpasses the previous state-of-the-art which takes multiple modalities including visual, auditory and textual inputs~\cite{kazakos2021little} by a more than $10\%$ absolute margin on the mean accuracy metric.
Since previous methods are based on different backbones, we experiment with a TSF-Base (``Visual only'') model pre-trained on Kinetics~\cite{carreira2017i3d} as a fair baseline for \ours.
We observe that its accuracy is comparable to previous methods but much lower than \ours, implying the effectiveness of learning visual representation on large-scale egocentric videos and using LLM as textual supervision during pre-training.

\myparagraph{\charadesego.} Next, we compare \ours's representation on the \charadesego action classification task. As shown in~\cref{tab:sota_charadesego}, \ours's representation excels on this task as well, which is notable as CharadesEgo videos are significantly different compared to Ego4D, being captured by crowdsourced workers using mobile cameras.

\subsection{Application to Third-Person Video Pre-training}
\label{sec:expt:third_person}

We apply \ours to third-person videos by experimenting with the HowTo100M~\cite{miech2019howto100m} dataset.
Specifically, we use the temporally aligned subset provided by~\cite{han2022tan}, which contains 3.3M sentences from 247k videos.
We evaluate the video representation on two third-person video datasets,~\ie UCF-101~\cite{soomro2012ucf101} and HMDB-51~\cite{kuehne2011hmdb} for action classification using the linear probing protocol.
For more details, please refer to \cref{sec:appdx:impl}.
From \cref{tab:sota_linear_probe}, we see that \ours outperforms previous methods such as MIL-NCE~\cite{miech2020milnce} and TAN~\cite{han2022tan} by a large margin.
Since we use a different backbone, we report a baseline without LLM and show that \ours indeed benefits from the language supervision.

\begin{table}
	\tablestyle{2pt}{1.05}
	\begin{tabular}{y{72}|x{36}|x{36}|x{36}H}
		Method & Vis. Enc. & UCF-101 & HMDB-51 & Kinetics-400 \\
		\thickhline		
		MIL-NCE~\cite{miech2020milnce}  & S3D & 82.7 & 54.3  & \\
		TAN~\cite{han2022tan} & S3D & 83.2 & 56.7 & \\
		\arrayrulecolor{LightGrayForTableRule}
		\hline
		\arrayrulecolor{Black}
		Baseline (w/o LLM) & TSF-B & \underline{86.5} & \textbf{59.4} & \\
		\ours & TSF-B & \textbf{87.4} & \underline{57.2} & \\
		\rowcolor{Gray} \ours & TSF-L & \textbf{88.1} & \textbf{61.5} & \\
		\hline
	\end{tabular}
	\caption{\textbf{\ours on third-person videos}. We measure the linear-probing action classification performance of the video model after pre-training on HowTo100M~\cite{miech2019howto100m}.
	}
	\label{tab:sota_linear_probe}
\end{table}

\begin{table*}[t]
	\centering
	\subfloat[
	\textbf{Generation Quality}.
	Using a sufficiently large language model as the text decoder is crucial for good text generation quality and downstream performance.
	\label{tab:vlm_gen_quality}
	]{
		\begin{minipage}{0.38\linewidth}{\begin{center}
					\tablestyle{1pt}{1.05}
					\begin{tabular}{HHx{32}x{30}x{24}|HHHHx{20}x{20}x{20}|x{24}H}
						Vis. Enc. arch. & Vis. Enc. init. & Text Dec. arch. & Text Dec. init. & Freeze LM & \rotatebox{45}{Bleu-1} & \rotatebox{45}{Bleu-2} & \rotatebox{45}{Bleu-3} & \rotatebox{45}{Bleu-4} &  \rotatebox[origin=t]{45}{\scriptsize{METEOR}} & \rotatebox[origin=t]{45}{\scriptsize{ROUGE-L}} & \rotatebox[origin=t]{45}{\scriptsize{CIDEr}} & Avg. mAP & Avg. nDCG \\
						\thickhline
						\multicolumn{5}{c|}{(baseline)} & - & - & - & - & - & - & - & 26.0 &  29.0 \\
						TSF-B & WIT  & GPT-2  & random &  \xmark & 0.507 & 0.394 & 0.306 & 0.200 & 0.284 & 0.524 & 0.882 & 24.3 & 28.1 \\
						TSF-B & WIT  & GPT-2  & WebText & \cmark & 0.483 & 0.386 & 0.300 & 0.191 & 0.270 & 0.505 & 0.804 & 24.0 & 28.1 \\
						TSF-B & WIT+Ego4D & \scriptsize{GPT-2 XL} & WebText & \cmark & {\bf 0.509} & {\bf 0.401} & {\bf 0.314} & {\bf 0.208} & {\bf 0.289} & {\bf 0.530} & {\bf 0.940} & {\bf 26.2} & {\bf 29.4} \\
						\hline
						\multicolumn{7}{c}{} \\
					\end{tabular}
			\end{center}}
		\end{minipage}
	}
	\hspace{0.8ex}
	\subfloat[
	\textbf{Sampling}.
	\ours benefits more from narrations produced by nucleus sampling than beam search.
	\label{tab:sampling}
	]{
		\begin{minipage}{0.22\linewidth}{\begin{center}
					\tablestyle{1pt}{1.05}
					\begin{tabular}{x{44}x{36}H|x{24}H}
						Sampling method & \# of sentences & filter & Avg. mAP & Avg. nDCG  \\
						\thickhline
						\scriptsize{N/A (baseline)}      & - & n/a & 26.0 & 29.0 \\
						\scriptsize{Beam search}  & 1 & \xmark &  27.9 & 30.2 \\
						Nucleus & 1 & \xmark & 29.6 & 31.8 \\
						Nucleus  & 10 & \xmark & {\bf 29.7} & {\bf 31.5} \\
						\hline
						\multicolumn{3}{c}{} \\
					\end{tabular}
		\end{center}}\end{minipage}
	}
	\hspace{0.8ex}
	\subfloat[
	\textbf{Scaling effect of \ours.}
	Gains increase on scaling the video encoder in \narrator. Default refers to only using  the original narrations.
	\label{fig:model_scaling}
	]{
		\resizebox{0.32\linewidth}{!}{
		\begin{tikzpicture}
	\begin{axis}[
		ybar=10pt,
		bar width=30pt,
		enlargelimits=0.15,
		axis y line*=left,
		axis x line*=bottom,
		legend style={at={(0.2,0.95)},
		anchor=north,legend columns=1},
		ylabel={Avg. mAP},
		xlabel={\narrator's architecture},
		symbolic x coords={Default, TSF-B, TSF-L, TSF-L@HR},
		xtick=data,
		nodes near coords,
		nodes near coords align={vertical},
		every node near coord/.append style={font=\Large, Black, /pgf/number format/.cd,fixed zerofill,precision=1},
		width=\linewidth,
		height=0.45\linewidth,
		ymin=25, ymax=35,
		ymajorgrids = true,
		ylabel style={font=\Huge},
		xlabel style={font=\Huge},
		tick label style={font=\huge},
		legend style={font=\huge, at={(0.03,0.85)}, anchor=west}, %
		legend cell align={left},
		]
		\addlegendimage{empty legend}
		\addplot [NiceBlue!50!white,fill=NiceBlue!50!white] coordinates {
		(Default, 26.0) (TSF-B, 26.2) (TSF-L, 28.1) (TSF-L@HR, 29.7)};
		\addplot [color_our!80!white,fill=color_our!80!white] coordinates {
		(Default, 29.8) (TSF-B, 29.6) (TSF-L,31.0) (TSF-L@HR,35.0)};
		\addlegendentry{\hspace{-0.2cm}Dual-Encoder's Video Architecture}
		\addlegendentry{TSF-B}
   		\addlegendentry{TSF-L}
	\end{axis}
\end{tikzpicture}%
		}
	}
	\vspace{-15pt}
	\caption{\textbf{Ablations of \narrator}. We report zero-shot average mAP on \ekmir for comparing downstream performance.
	We study \narrator from the perspective of generation quality ({\em left}), sampling techniques ({\em middle}), and scaling effect ({\em right}).
	}
	\label{tab:ablations_llm}
\end{table*}

\begin{table}[t]
	\tablestyle{1.9pt}{1.05}
	\begin{tabular}{x{20}x{20}x{24}|x{24}x{25}|x{24}x{24}|x{20}x{20}}
		\multirow{2}{*}{Rephr.} &   \multirow{2}{*}{Recap.}   &    Pseudo    & \multicolumn{2}{c|}{\ekmir} & \multicolumn{2}{c|}{\egomcq}       & \multicolumn{2}{c}{EGTEA} \\
		&       &  cap.   & {\tiny Avg. mAP}      & {\tiny Avg. nDCG}    & {\tiny inter-video} & {\tiny Intra-video}  & Mean         & Top-1          \\
		\thickhline
		&                &             &     26.0   &     28.8    &      {\bf 93.6}        &   54.3     &      27.3          &      30.1           \\
		\cmark &                &               &     28.0  &     30.1     &     93.5      &   56.9      &       \underline{29.8}         &      30.8         \\
		&   \cmark   &            &   27.1    &   29.9   &   93.2     &   {\bf 59.2}    &  26.8  & 31.2   \\
		\cmark &   \cmark  &              &  \underline{29.7}    &   {\bf 31.5}   &  {\bf 93.6}  &   58.3    &   29.4    &     {\bf 36.6}   \\
		\cmark &   \cmark   &    \cmark    &   {\bf 29.9}    &   \underline{31.4}    &  {\bf 93.6}  &   \underline{59.1}  &    {\bf 31.1} &  \underline{36.0}  \\
		\hline
	\end{tabular}
	\caption{\textbf{Contributions of different Language Supervision.}
	We can see that (1) using \rephraser (``Rephr.'') and \narrator (``Recap.'') improve downstream zero-shot performance complementarily, (2) dense pseudo-captioning further improves performance on 3 out of 6 metrics.
	}
	\label{tab:exp:ablation_timesformer_B}

\end{table}

\subsection{Application to Semi-supervised Learning}
\label{sec:expt:semi_sup}

While \ours is very effective at leveraging existing narrations to %
augment them, we now show that it is also applicable when only a limited number of narrations are available to begin with.
We first divide each long video from Ego4D into 15-second chunks and assume only the annotated clips within every $N$ chunks is available during pre-training, leading to approximately $\frac{100}{N}\%$ of the full set, where $N\in\{2,5, 10\}$.
This can be considered a practical scenario when we want to annotate as many videos as possible for diversity when the annotation budget is limited.
In the remainder $(1-\frac{100}{N}\%)$ part that is skipped, we uniformly sample the same number of the clips per chunk with the same clip length as that in the seen chunks.
Both the dual-encoder model and \narrator are trained on the $\frac{100}{N}\%$ available annotations.

We plot the zero-shot performance curve of pre-training with different proportions in \cref{fig:percentage}.
We can see that \ours consistently outperforms the ground-truth-only baseline at all points (10, 20, 50, and 100\%).
The improvement tends to be larger when more data is available, indicating the method's scalability as more videos are narrated in the future.
Furthermore, we observe our method can achieve a similar level of performance with the baseline often using less than 50\% data.
We also achieve a comparable result with the state-of-the-art using much fewer data.

\subsection{Ablation Studies}
\label{sec:expt:ablation}

\myparagraph{Contributions of Different Language Supervisions}.
We ablate different language supervisions in~\cref{tab:exp:ablation_timesformer_B} on \ek MIR (zero-shot), EgoMCQ and EGTEA.
Using the text-only \rephraser (``rephr.'') or visually conditioned \narrator (``recap.'') separately improves the ground-truth baseline noticeably.
Combining both \rephraser and \narrator gives an improvement of 3.5\% average mAP on EK-100 MIR.
We see that dense captioning on the entire video (``pseudo-cap.'') is also helpful. Though the gain on \ekmir is not as significant, it shows nontrivial improvements on \egomcq intra-video accuracy and EGTEA mean accuracy.
Our conjecture for this marginal gain is that informative clips are mostly covered in Ego4D because all videos are inspected by two annotators.

\myparagraph{Generation Quality of \narrator}.
We study how the \narrator's configurations affect the quality of generated text and the downstream performance.
The generation quality is measured by standard unsupervised automatic metrics including METEOR, ROUGE, and CIDEr~\cite{maluuba2022nlgeval}.
We use a \narrator with a smaller GPT-2 as the text decoder and consider two scenarios in~\cref{tab:vlm_gen_quality}: (1) LM is randomly initialized but jointly trained with the gated cross-attention modules, and (2) LM is initialized from the original GPT-2.
The generation quality decreases compared to GPT-2 XL in both cases and the zero-shot retrieval result on \ekmir is worse.
This indicates that the language model should be sufficiently large and pre-trained on web text data.

\myparagraph{Sampling}.
In~\cref{tab:sampling}, we investigate different sampling methods for text generation from \narrator.
We see that nucleus sampling works much better than beam search while repetitive sampling shows marginal improvement.

\myparagraph{Scaling effect}.
In~\cref{fig:model_scaling}, we compare the zero-shot retrieval result by progressively increasing the size of \narrator's video encoder from TSF-B to TSF-L and TSF-L@HR, which increases the input resolution to be narrated from 224 to 336 while fixing the dual-encoder architecture.
The retrieval performance steadily increases while \narrator becomes stronger.
We conduct this experiment by varying the dual-encoder architecture, namely TSF-Base and TSF-Large, and show similar trends.
Both phenomena suggest that \ours can scale to larger models.

\section{Conclusion and Future Work}

In this paper, we proposed \ours, a new approach to video-language representation learning by automatically narrating long videos with LLMs. We achieve strong improvements over baselines trained with the same amount of human-narrated videos and set new state-of-the-art on six popular benchmark tasks across first- and third-person video understanding benchmarks. \ours also shows positive scaling behavior when adding more training narrations, using larger visual backbones, and using stronger LLMs, all of which are promising areas for future work.

{\noindent \bf Acknowledgements:}
We thank Naman Goyal, Stephen Roller and Susan Zhang for help with language models,
Kevin Qinghong Lin for help with EgoVLP, and the Meta AI team for helpful discussions and feedback.
This material is based upon work in-part supported by the National Science Foundation under Grant No. IIS-1845485.

{\small
\bibliographystyle{ieee_fullname}
\bibliography{refs}
}

\clearpage
\appendix

\section{Radar Chart \Cref{fig:teaser_sota} Details}
\label{sec:appdx:radar_chart}
We first describe how we plot the radar chart in \cref{fig:teaser_sota}.
Each axis denotes a specific metric on one video understanding task.
Each vertex denotes a ratio relative to our performance, which is computed by normalizing the performance of either \ours or previous SOTA by that of \ours, and is in the range of $(0, 1]$.
For illustrative purpose, we set the radar chart's origin to be 80\% and outermost frame to be 100\% so that the interval between neighboring lattices to be 5\%.
The numbers annotated next to the vertices are {\em absolute value} of performance {\em without normalization}.
Note that in other radar charts~\cite{yu2022coca,wang2022beitv3}, the axes have different scales and interval values while the origin is not valid, which may lead to potential fallacies.

\section{ \bf \ours Details}
\label{sec:appdx:method}

The algorithm of training \ours is given in \cref{alg:lavie}.
The loss is based on the CLIP~\cite{radford2021clip}'s symmetric cross-entropy loss over the similarity scores of samples in the batch $  \mathcal{\widetilde{B}}_l \cup  \mathcal{\widetilde{B}}_u $ with minimal modifications.
We apply two separate temperatures $(\tau_{r}, \tau_{n})$ for embeddings from rephrased pairs and pseudo-captioned ones respectively,
{\small
	\begin{align}
		\label{eq:loss}
		\mathcal{L}= -\frac{1}{2N} \sum_{i=1}^N \left(
		\log{\frac{\exp(\frac{\vv_i^\top\vu_i}{\tau_i})}{\sum\limits_{j=1}^N \exp(\frac{\vv_i^\top\vu_j}{\sqrt{\tau_i\tau_j}})}} + \log{\frac{\exp(\frac{\vu_i^\top\vv_i}{\tau_i})}{\sum\limits_{j=1}^N \exp(\frac{\vu_i^\top\vv_j}{\sqrt{\tau_i\tau_j}})}}
		\right).
	\end{align}
}
We ablate different choices of temperatures in \cref{tab:temperature}.

\section{Dataset Details}\label{sec:appdx:dataset}

In this section, we provide details of the datasets where we conduct experiments.

\myparagraph{Ego4D}.
Ego4D contains 3,670 hours of egocentric videos with temporally dense narrations.
Each narration has a timestamp and an associated free-form sentence.
We construct the video-text clip pairs that are used for pre-training following~\cite{lin2022egovlp}.
First, we exclude 2,429 videos that appear in the validation and test sets of the Ego4D benchmark.
Next, we determine the each clip's interval using the contextual variable-length clip pairing strategy in~\cite{lin2022egovlp}.
Finally, we drop the narrations that either contain ``\#unsure''/``\#Unsure'' tags or are shorter than 4 words.
This results in 4,012,853 video-text clip pairs with an average clip length of $1 (\pm0.9)$ second.
For the excluded videos, we also pre-process similarly and obtain 1,260,434 video-text clip pairs.
We only use them as validation split to measure the generation quality of \narrator in \cref{tab:vlm_gen_quality}.

\begin{algorithm}[t]
	\caption{One step of training \ours}
	\label{alg:lavie}
	\begin{algorithmic}
		\Require A subset of narrated (unnarrated) clips $\mathcal{B}_l$ ($\mathcal{B}_u$)
		\State clips with LM-generated narrations $ \mathcal{\widetilde{B}}_l = \{\},  \mathcal{\widetilde{B}}_u = \{\} $
		\For{$(x_i, y_i)\in\mathcal{B}_l$}
		\State $ u \sim U(0, 1) $ \Comment{Uniform sample between 0 and 1}
		\If{$u<0.5$} \Comment{Query \rephraser}
		\State $y'_i \sim p_\mathrm{\rephraser}(y'|y_i), \tau_i \gets \tau_{r}$
		\Else  \Comment{Query \narrator}
		\State $y'_i \sim p_\mathrm{\narrator}(y'|x_i), \tau_i \gets \tau_{n}$
		\EndIf
		\State $ \mathcal{\widetilde{B}}_l \gets \mathcal{\widetilde{B}}_l \cup \{ (x_i, y'_i, \tau_i) \}$
		\EndFor
		\For{$x_i\in\mathcal{B}_u$}
		\State $y'_j \sim p_\mathrm{\narrator}(y'|x_i), \tau_j \gets \tau_{n}$
		\State $ \mathcal{\widetilde{B}}_u \gets \mathcal{\widetilde{B}}_u \cup \{ (x_j, y'_j, \tau_j) \}$
		\EndFor
		\State Train $\mathcal{F}_\mathrm{\ours}(x, y)$ with the batch $  \mathcal{\widetilde{B}}_l \cup  \mathcal{\widetilde{B}}_u $ using \Cref{eq:loss}.
	\end{algorithmic}
\end{algorithm}

\myparagraph{\ek}.
The Epic-Kitchens-100 (\ek) dataset contains 100 hours of egocentric cooking videos.
The training split has 67,217 video clips; the validation split has 9,668 video clips; the testing split has 13,092 video clips.
Each clip is annotated with (1) a start and end timestamp, (2) a short textual narration, and (3) a verb and noun class that the narration belongs to.
The action class can also be uniquely determined by combining the verb and the noun.
In the zero-shot setting, we evaluate the pre-trained model on the validation split directly without any tuning;
In the finetuned setting, we take the pre-trained model and perform end-to-end finetuning on the training split and evaluate on the validation split.
For \ekmir we use the textual narration while for \ekcls we use the class of verb, noun, and action as the label.
For \ekmir, the evluation metrics are mean Average Precision (mAP) and normalized Discounted Cumulative Gain (nDCG).
For \ekcls, the evaluation metrics are top-1 accuracies for verb, noun, and action.
Action-level accuracy is the most important one among all.

\myparagraph{EGTEA}.
EGTEA contains 28 hours of egocentric cooking videos with gazing tracking.
In our experiments, we take as input the visual frames only.
The action annotations include 10,321 instances of fine-grained actions from 106 classes, with an average duration of 3.2 seconds.
In the zero-shot setting, we evaluate the pre-trained model on the test set of all three splits without any tuning and report results as the mean accuracy averaged across all classes across all three splits, as Li~\etal~\cite{li2018egtea} suggested.
In the finetuned setting, we follow prior works~\cite{kazakos2021little} and report top-1 accuracy and mean class accuracy using the first train/test split, which has 8,299/2,022 instances respectively.

\myparagraph{CharadesEgo.}
The CharadesEgo dataset contains 7,860 videos of daily indoor activities from both third- and first-person views.
The annotations are 68,536 instances of fine-grained actions from 157 classes.
We use the first-person subset only, comprising 3,085 videos for training and 846 videos for testing.
We report video-level mAP as the evaluation metric.
In the zero-shot setting, we evaluate the pre-trained model on the test videos directly without any tuning;
In the finetuned setting, we perform end-to-end finetuning on the trimmed action instances in the training split, which has an amount of 33,114 action instances.

\section{Implementation Details}
\label{sec:appdx:impl}

\subsection{Pre-training on Ego4D}
\label{sec:appdx:impl:pretrain_ego4d}
We pre-train on the video-narration pairs from Ego4D~\cite{grauman2022ego4d}.
We train the model using AdamW with $(\beta_1,\beta_2)=(0.9, 0.999)$ and weight decay of 0.01 for 5 epochs.
We use a fixed learning rate of 3e-5.
The projection head after the dual-encoders is a linear layer with an output dimension of 256.
We use PyTorch's native FP16 mixed precision training and gradient checkpoint.
This allows us to afford a per-gpu batch size of 32 over 32 GPUs for TimeSformer-B and a per-gpu batch size of 16 over 64 GPUs for TimeSformer-L, resulting in a total batch size of 1,024.
We abate these design choices in \cref{sec:appdx:ablations}.

For input, we first divide each video into 5-minute segments and scale the short side of the video to 288 pixels.
This signifantly reduces storage and accelerates decoding.
During training, we decode the corresponding segment that contains the selected clip.
We randomly sample 4 frames between the start and end time of the clip and use standard \texttt{RandomResizedCrop (0.5, 1.0)} for data augmentation.

\subsection{Training \narrator on Ego4D}
\myparagraph{Architecture.}
For the video encoder, we use the one we obtain in \cref{sec:appdx:impl:pretrain_ego4d} and keep it frozen.
We drop the global average pooling layer and attach an attention pooling module, which is instantiated by a standard cross-attention~\cite{vaswani2017attention} and a Layer Normalization~\cite{ba2016layernorm}.
The attention pooling uses a fixed length of randomly initalized queries $\mathbf{q}\in\mathbb{R}^{N_\mathrm{q}\times D_\mathrm{t}}$ to attend visual features $\mathbf{v}\in\mathbb{R}^{(T\times H' \times W')\times D_\mathrm{v}}$.
This results in a fixed length of hidden states, $\mathrm{AttentionPool}(\mathbf{q}, \mathbf{v})\in\mathbb{R}^{N_\mathrm{q}\times D_\mathrm{t}}$, which will be later fed into the cross-attention module of the text decoder. 
This ensures the text decoder attends to the same number of visual features irrespective of the input visual resolution, \eg 224$\times$224 or 336$\times$336.
More concretely, $ \mathrm{AttentionPool}(\mathbf{q}, \mathbf{v}) $ is computed as follows:
{\small
\begin{align*}
\mathbf{q}', \mathbf{v}' &= \mathrm{LayerNorm}(\mathbf{q}), \mathrm{LayerNorm}(\mathbf{v}), \\
\mathrm{head}_i &= \mathrm{softmax}\left(\frac{(\mathbf{q}'\mathbf{W}_Q^{(i)}) (\mathbf{v}'\mathbf{W}_K)^\top}{\sqrt{d_0}}\right)\cdot(\mathbf{v}'\mathbf{W}_V), \\
\mathrm{AttentionPool} &= \mathrm{Concat}(\mathrm{head}_1, \cdots, \mathrm{head}_h)\cdot\mathbf{W}_O,
\end{align*}
}
where $\mathbf{W}_Q\in\mathbb{R}^{D_\mathrm{t} \times d_0}$, $\mathbf{W}_{K/V}\in\mathbb{R}^{D_\mathrm{v} \times d_0}$, and $\mathbf{W}_O\in\mathbb{R}^{(h\cdot d_0) \times D_\mathrm{t}}$.

For the text decoder, we use GPT-2 XL~\cite{radford2019gpt2} and keep it frozen.
The video encoder and the text decoder is bridged by a cross-attention module.
Each cross-attention module comprises a cross-attention layer followed by a feed-forward network (FFN).
Layer Normalization is added at the beginning of both cross-attention and FFN.
We add \texttt{tanh}-gating~\cite{hochreiter1997lstm} with an initial value of zero.
We insert one cross-attention module every two GPT2-Blocks in GPT2 XL to save memory.
Both the attention pooling and cross-attention modules are learnable parameters.

We train \narrator on the ground-truth video-narration pairs from Ego4D~\cite{grauman2022ego4d}.
The training recipe mostly follows the one for pre-training the dual-encoders except that we use FP32 to train \narrator because PyTorch's native FP16 mixed-precision leads to training instability.
We use the video-text clip pairs from the Ego4D's validation videos to compute the word-level classification accuracy and perplexity.
We select the model with the highest accuracy as well as lowest perplexity, which is often reached after 3$\sim$4 epochs.

\subsection{Multi-Instance Retrieval on EK-100}
We fine-tune the pre-trained model on EK100 using AdamW with $(\beta_1,\beta_2)=(0.9, 0.999)$ and weight decay of 0.01.
We use cosine annealing with warmup, where the base learning rate starts from 1e-6, linearly increases to a peak of 3e-3 in the first epoch and then gradually decreases to 1e-5 following a half-wave cosine schedule.
We apply the multi-instance max-margin loss~\cite{wray2019jpose} with a margin value of 0.2.
We use a per-gpu batch size of 16 over 8 GPUs for TimeSformer-B and a per-gpu batch size of 4 over 32 GPUs for TimeSformer-L.
We use a stochastic depth ratio of 0.1 in the backbone.

For the input, we represent each video clip with 16 sampled frames at both training and testing time.
At training time, We scale the short side of the video to 256 pixels and then take a 224$\times$224 crop while at testing time, we scale the short side to 224 pixels and take the center 224$\times$224 crop.

\subsection{Action Recognition on EGTEA}
We fine-tune the pre-trained model on EGTEA for 100 epochs using SGD with a momentum of 0.9 and weight decay of 5e-4.
We use cosine annealing with warmup, where the base learning rate starts from 1e-6, linearly increases to a peak of 3e-3 in the first epoch and then gradually decreases to 1e-5 following a half-wave cosine schedule.
We drop the linear projection head and attach a $106$-dim head for classification.
For \ours, we train the classification head with $1\times$ base learning rate and the backbone with $0.1\times$.
For visual-only video model pre-trained on Kinetics, we use $1\times$ base learning rate  for both the classification head and the backbone.
We use a per-gpu batch size of 16 over 8 GPUs for TimeSformer-B and a per-gpu batch size of 4 over 32 GPUs for TimeSformer-L.
We use a stochastic depth ratio of 0.1 in the backbone and a dropout of 0.5 before the classification head.
We also use a label smoothing of 0.1.

For input, we randomly select a 32-frame video clip at a temporal stride of 2 (namely 16$\times$2) from each video at training time.
We scale the short side of the video to 256 pixels and then take a 224$\times$224 crop.
For data augmentation, we use standard \texttt{RandomResizedCrop (0.5, 1.0)} and \texttt{RandomHorizontalFlip(0.5)}.
At testing time, we evenly take ten 32-frame clips through the full video.
We scale the short side to 224 pixels and take three spatial crops along the longer axis per clip.
The final predictions are averaged over all these crops.

\subsection{Action Recognition on EK-100}
We fine-tune the pre-trained model on EK100 with a same training schedule as in EGTEA.
The only exception is that we apply three classification heads for verb, noun, and action separately because we empirically observe that it speeds up convergence and performs slightly better than using a single action-level classification head.

For the input, we represent each video clip with 16 sampled frames at both training and testing time.
At testing time, we take three spatial crops along the longer axis per clip and average the final predictions.

\begin{table}
	\tablestyle{2pt}{1.05}
	\begin{tabular}{y{96}|x{60}|x{20}x{20}x{20}}
		\multirow{2}{*}{Method (Backbone)}&  \multirow{2}{*}{Pretrain} & \multicolumn{3}{c}{Top-1 accuracy} \\
		&  & Verb & Noun & Action \\
		\thickhline
		IPL (I3D)~\cite{wang2021ipl} & K400 & 68.6 & 51.2 & 41.0 \\
		ViViT-L~\cite{arnab2021vivit} & IN-21k+K400 & 66.4 & 56.8 & 44.0 \\
		MoViNet~\cite{kondratyuk2021movinets} & N/A & \textbf{72.2} & 57.3 & 47.7 \\
		MTV~\cite{yan2022multiview}  & WTS-60M & 69.9 & \textbf{63.9} & \underline{50.5} \\
		MTCN (MFormer-HR)~\cite{kazakos2021little} & {\tiny IN-21k+K400 +VGG-Sound} & 70.7 & 62.1 & 49.6 \\
		Omnivore (Swin-B)~\cite{girdhar2022omnivore} & {\tiny IN21k+IN-1k +K400+SUN} & 69.5 & 61.7 & 49.9 \\
		MeMViT~\cite{wu2022memvit}  & K600 & 71.4 & 60.3 &  48.4 \\
		\arrayrulecolor{LightGrayForTableRule}
		\hline
		\ours (TSF-L) & WIT+Ego4D & \underline{72.0} & \underline{62.9} & \textbf{51.0} \\
		\arrayrulecolor{Black}
		\hline
	\end{tabular}
	\caption{\textbf{The performance of action recognition on \ek}. We report top-1 accuracy on verb, noun, and action. \ours outperforms all prior works in terms of action-level top-1 accuracy.}
	\label{tab:sota_ek100_cls}
\end{table}

\subsection{Action Recognition on CharadesEgo}
Following EgoVLP~\cite{lin2022egovlp}, we convert the task of action classification to that of video-text retrieval as follows:
for each trimmed video clip with textual annotations, we consider it to be a valid video-text pair for training.
Since CharadesEgo is a multi-class dataset, which means each trimmed video can be annotated with different classes, we treat any trimmed video clip with $N$ actions as $N$ individual video-text pairs.
We use the same InfoNCE~\cite{oord2018cpc} loss.
We fine-tune the pre-trained model on CharadesEgo using AdamW with $(\beta_1,\beta_2)=(0.9, 0.999)$ and weight decay of 0.01.
We use cosine annealing with warmup, where the peak learning rate is set to be 3e-5.
For input, we randomly select a 32-frame video clip at a stride of 2 from the
\emph{trimmed} video at training time and evenly sample 16 frames from the \emph{untrimmed} video at testing time to calculate the video-level mAP.
We finetune the model for 10 epochs and report the best performance.

\subsection{\ours for Third-person Video Pre-training}
The pre-training recipe mostly follows the one in \cref{sec:appdx:impl:pretrain_ego4d} except that when constructing a batch of samples, we sample one more hard negative clip from the same video  for each selected clip following~\cite{han2022tan}.

When doing linear-probing evaluation, we keep the video encoder frozen, extract video feature and train a linear SVM on top.
For each video clip in either HMDB-51 or UCF-101, we evenly take four 32-frame clips through the entire video.
We scale the short side to 224 pixels and take the center crop per clip and pass through the frozen video encoder to get the final visual embedding.
For each testing video, we average the prediction score from different clips.
We use scikit-learn's LinearSVC and report the highest top-1 accuracy after sweeping the regularization parameter $C\in\{10^{-5}, 10^{-4}, 10^{-3}, 10^{-2}, 0.1, 1, 10^2, 10^3, 10^4\}$.

\section{Additional Results}\label{sec:appdx:results}

\myparagraph{\ekcls}.
We compare \ours representation on \ekcls in \cref{tab:sota_ek100_cls}.
We achieve state-of-the-art performance in terms of top-1 action accuracy.
Note that the second best-performing Multiview Transformer~\cite{yan2022multiview} is pre-trained on WTS-60M which is not publicly available.

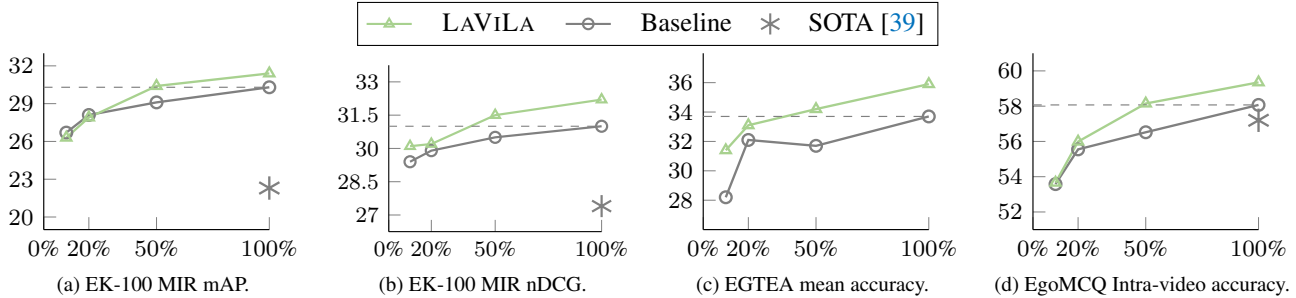
\begin{figure*}[t]
	\centering
	\begin{subfigure}[b]{\linewidth}
		\centering
		\begin{tikzpicture}
			\begin{customlegend}
				[
				legend columns=3, legend style={column sep=2ex},
				legend entries={\ours,Baseline,SOTA~\cite{lin2022egovlp}
				}]
				\addlegendimage{mark=triangle,style={thick},color_our}
				\addlegendimage{mark=o,style={thick},gray}
				\addlegendimage{mark=asterisk,color_ao,mark options={scale=2,thick},only marks}
			\end{customlegend}
		\end{tikzpicture}
	\end{subfigure}
	
	\begin{subfigure}[b]{0.246\linewidth}
		\resizebox{\textwidth}{!}{
			\begin{tikzpicture}
	\begin{axis} [
		axis x line*=bottom,
		axis y line*=left,
		legend pos=north east,
		ymin=19, ymax=33,
		xmin=0, xmax=100,
		xticklabel={\pgfmathparse{\tick}\pgfmathprintnumber{\pgfmathresult}\%},
		xtick={0,20,50,100},
		ytick={20,23,26,29, 32},
		width=\linewidth,
		legend style={cells={align=left}},
		label style={font=\footnotesize},
		tick label style={font=\footnotesize},
		legend style={at={(0.35,0.2)},anchor=west},
		]
		
		\addplot[mark=o,style={thick},gray] plot coordinates {
			(10, 26.7)
			(20, 28.1)
			(50, 29.1)
			(100, 30.3)
		};
		\addplot[mark=triangle,style={thick},color_our] plot coordinates {
			(10, 26.3)
			(20, 27.9)
			(50, 30.4)
			(100, 31.4)
		};
		\addplot[dashed,gray] plot coordinates {
			(0, 30.3)
			(100, 30.3)
		};
		\addplot[mark=asterisk,color_ao,mark options={scale=2, thick},only marks] plot coordinates {
			(100, 22.3)
		};
		\label{pgf:ek100_map}
	\end{axis}
\end{tikzpicture}%
		}
		\caption{\ekmir mAP.}
	\end{subfigure}
	\begin{subfigure}[b]{0.246\linewidth}
		\resizebox{\textwidth}{!}{
			\begin{tikzpicture}
	\begin{axis} [
		axis x line*=bottom,
		axis y line*=left,
		legend pos=north east,
		ymin=26.25, ymax=33.75,
		xmin=0, xmax=100,
		xticklabel={\pgfmathparse{\tick}\pgfmathprintnumber{\pgfmathresult}\%},
		xtick={0,20,50,100},
		ytick={27,28.5,30,31.5,33},
		width=\linewidth,
		legend style={cells={align=left}},
		label style={font=\footnotesize},
		tick label style={font=\footnotesize},
		legend style={at={(0.35,0.2)},anchor=west},
		]
		
		\addplot[mark=o,style={thick},gray] plot coordinates {
			(10, 29.4)
			(20, 29.9)
			(50, 30.5)
			(100, 31.0)
		};
		\addplot[mark=triangle,style={thick},color_our] plot coordinates {
			(10, 30.1)
			(20, 30.2)
			(50, 31.5)
			(100, 32.2)
		};
		\addplot[dashed,gray] plot coordinates {
			(0, 31.0)
			(100, 31.0)
		};
		\addplot[mark=asterisk,color_ao,mark options={scale=2, thick},only marks] plot coordinates {
			(100, 27.4)
		};
		\label{pgf:ek100_ndcg}
	\end{axis}
\end{tikzpicture}%
		}
		\caption{\ekmir nDCG.}
	\end{subfigure}
	\begin{subfigure}[b]{0.246\linewidth}
		\resizebox{\textwidth}{!}{
			\begin{tikzpicture}
	\begin{axis} [
		axis x line*=bottom,
		axis y line*=left,
		legend pos=north east,
		ymin=26, ymax=38,
		xmin=0, xmax=100,
		xticklabel={\pgfmathparse{\tick}\pgfmathprintnumber{\pgfmathresult}\%},
		ytick={28,30,32,34,36},
		xtick={0,20,50,100},
		width=\linewidth,
		legend style={cells={align=left}},
		label style={font=\footnotesize},
		tick label style={font=\footnotesize},
		legend style={at={(0.35,0.2)},anchor=west},
		title style={font=\Large},
		]
		\addplot[mark=o,style={thick},gray] plot coordinates {
			(10, 28.2)
			(20, 32.1)
			(50, 31.7)
			(100, 33.7)
		};
		\addplot[mark=triangle,style={thick},color_our] plot coordinates {
			(10, 31.4)
			(20, 33.1)
			(50, 34.2)
			(100, 35.9)
		};
		\addplot[dashed,gray] plot coordinates {
			(0, 33.7)
			(100, 33.7)
		};
		\label{pgf:egtea_mean}
	\end{axis}
\end{tikzpicture}%
		}
		\caption{EGTEA mean accuracy.}
	\end{subfigure}
	\begin{subfigure}[b]{0.246\linewidth}
		\resizebox{\textwidth}{!}{
			\begin{tikzpicture}
	\begin{axis} [
		axis x line*=bottom,
		axis y line*=left,
		legend pos=north east,
		ymin=51, ymax=61,
		xmin=0, xmax=100,
		xticklabel={\pgfmathparse{\tick}\pgfmathprintnumber{\pgfmathresult}\%},
		xtick={0,20,50,100},
		ytick={52,54,56,58,60},
		width=\linewidth,
		legend style={cells={align=left}},
		label style={font=\footnotesize},
		tick label style={font=\footnotesize},
		legend style={at={(0.35,0.2)},anchor=west},
		title style={font=\Large},
		]
		\addplot[mark=o,style={thick},gray] plot coordinates {
			(10, 53.58)
			(20, 55.55)
			(50, 56.52)
			(100, 58.07)
		};
		\addplot[mark=triangle,style={thick},color_our] plot coordinates {
			(10, 53.66)
			(20, 55.99)
			(50, 58.15)
			(100, 59.35)
		};
		\addplot[dashed,gray] plot coordinates {
			(0, 58.07)
			(100, 58.07)
		};
		\addplot[mark=asterisk,color_ao,mark options={scale=2, thick},only marks] plot coordinates {
			(100, 57.2)
		};
		\label{pgf:ego4d_mcq}
	\end{axis}
\end{tikzpicture}%
		}
		\caption{\egomcq Intra-video accuracy.}
	\end{subfigure}
	\caption{\textbf{More results of \ours in a semi-supervised setting where only a limited amount of narrations are given}. Both \ours and the baseline use a TimeSformer-Large as the visual encoder backbone. Comparing zero-shot performance of pre-training, \ours consistently outperforms the groundtruth-only baseline when 10, 20, 50,100\% data is used.}
	\label{fig:percentage_vitl}
\end{figure*}

\myparagraph{More results on Semi-supervised Learning}.
Following the setup in \cref{sec:expt:semi_sup}, we provide more results in \cref{fig:percentage_vitl} while replacing the backbone of \ours with TimeSformer-Large.
We observe similar trends as \cref{sec:expt:semi_sup} where \ours outperforms the ground-truth-only baseline at all data points.

\begin{table}[t]
	\tablestyle{2pt}{1.05}
	\begin{tabular}{y{48}|x{36}x{36}x{28}|x{28}x{28}}
		& EgoNCE  & CLIP-init. & \# frames & Avg. mAP & Avg. nDCG \\
		\thickhline
		\multicolumn{6}{c}{Extracted RGB frames}  \\
		\hline
		EgoVLP~\cite{lin2022egovlp} &  &  & 4 & 15.5 & 22.1  \\
		EgoVLP~\cite{lin2022egovlp} & \cmark & & 4 & 16.6 & 23.1  \\
		\hline
		\multicolumn{6}{c}{Videos (downsized to 480p)}  \\
		\hline
		EgoVLP~\cite{lin2022egovlp} & \cmark & & 4 & 22.3 & 27.4  \\
		EgoVLP~\cite{lin2022egovlp} & \cmark & & 16 & 23.6 & 27.9  \\
		Our impl.  &  &  & 4 & 24.1 & 28.0 \\
		Our impl.  &  & \cmark & 4 & 24.7 & 28.4 \\
		\hline
	\end{tabular}
	\caption{\textbf{Improved baseline} evaluted on \ekmir. We observe that evaluting on videos directly improves the baseline noticeably. Using CLIP-pre-trained encoder weights introduces additional improvements. All gains shown in the paper are on top of this already strong baseline (last row).}
	\label{tab:improved_baseline}
\end{table}

\begin{table*}[t]
	\centering
	\subfloat[
	\textbf{initialization}.
	IN-21K and WIT denote ImageNet-21k~\cite{deng2009imagenet} and WebImageText~\cite{radford2021clip}.
	BC+Wiki denotes BookCorpus+English Wikipedia on which BERT is pre-trained. Using CLIP-initialized weights works better than using those supervised pretrained on IK-21K. 
	\label{tab:weight_init}
	]{
		\centering
		\begin{minipage}{0.45\linewidth}{\begin{center}
					\tablestyle{4pt}{1.05}
					\begin{tabular}{x{36}x{30}x{36}x{30}|x{20}x{20}}
						Vis. Enc. arch. & Vis. Enc. init. & Text Enc. & Text Enc. init. & avg mAP & avg. nDCG  \\
						\thickhline
						TSF-B & IN-21K & DistilBERT & BC+Wiki & 24.1 & 28.0 \\
						TSF-B & WIT  & DistilBERT & BC+Wiki  & 24.2 &  {\bf 28.5} \\
						ViT-B  & WIT  & CLIP-GPT & WIT & 23.2 & 27.4 \\
						\rowcolor{Gray} TSF-B & WIT  & CLIP-GPT & WIT & {\bf 24.7} & 28.4 \\
						\hline
					\end{tabular}
		\end{center}}\end{minipage}
	}
	\hspace{0.5em}
	\subfloat[
	\textbf{Batch size}.
	Zero-shot performance improves when batch size increases from 512 to 1,024.
	\label{tab:batch_size}
	]{
		\begin{minipage}{0.23\linewidth}{\begin{center}
					\tablestyle{4pt}{1.05}
					\begin{tabular}{x{18}|x{24}x{24}}
						Batch size & Avg. mAP & Avg. nDCG \\
						\thickhline
						\rowcolor{Gray} 512 & 24.7 &  28.4 \\
						1024 & {\bf 25.6} & {\bf 28.8}  \\
						2048 & {\bf 25.6} & 28.5  \\
						\hline
					\end{tabular}
		\end{center}}\end{minipage}
	}
	\hspace{0.5em}
	\subfloat[
	\textbf{Projection head}.
	Zero-shot performance is affected by the hidden dimension of the projection head. Empirically using 256 yields a best performance.
	\label{tab:projection_head}
	]{
		\begin{minipage}{0.25\linewidth}{\begin{center}
					\tablestyle{1pt}{1.05}
					\begin{tabular}{x{48}|x{24}x{24}}
						Projection head & Avg. mAP & Avg. nDCG  \\
						\thickhline
						Linear ($128$-d) & 24.1 & 27.8  \\
						\rowcolor{Gray} Linear ($256$-d) & {\bf 24.7} & {\bf 28.4} \\
						Linear ($512$-d) &  24.5 & 28.1 \\
						\hline
					\end{tabular}
		\end{center}}\end{minipage}
	}
	\caption{\textbf{Ablations of dual-encoder}. We study how weight initialization (a), pre-training batch size (b), and project head dimension (c) affect the zero-shot performance of the dual-encoder on \ekmir.}
	\label{tab:ablations_vlp}
\end{table*}

\section{Additional Ablations}\label{sec:appdx:ablations}

\myparagraph{Improved Baseline on EK-100 MIR.}
We present an improved baseline of video-language model pretrained on Ego4D and evaluate it on EK-100 MIR in a zero-shot setting in \cref{tab:improved_baseline}.
The initial baseline is video-language model with a TimeSformer-Base as visual encoder and a Distil-BERT as textual encoder, proposed in EgoVLP~\cite{lin2022egovlp}.
First, we find that zero-shot evaluation on videos brings a noticeable improvement than on extracted RGB frames.
Particularly, given the same EgoVLP+EgoNCE model, zero-shot retrieval can increase by 5.7\% average mAP and 4.3\% average nDCG repespectively.
This is probably because frame extraction using ffmpeg's default parameter downgrades the image quality by a considerable amount.
Second, under the same video-as-input evaluation protocol, our implementation with the same backbone (TimeSformer-Base + DistilBERT) using standard InfoNCE loss \emph{without} EgoNCE, can achieve 24.1\% and 28.0\% average mAP and nDCG, better than the EgoVLP with EgoNCE.
Third, if we pretrain the joint model using CLIP-pretrained models as the initial weights, the zero-shot retrieval result can be further boosted (+0.6\% avg. mAP and +0.4\% avg. nDCG), indicating that egocentric video representation can also benefit from large-scale image-text pre-training.

Starting from this improved baseline, we conduct more ablations on pretraining the video-langauge model in \cref{tab:ablations_vlp} as follows.
We measure the performance by zero-shot average mAP and average nDCG on \ekmir.

\myparagraph{Effect of weight initialization}.
We study the effect of architectures and weight initialization in \cref{tab:weight_init}.
First, we observe that using the same architecture of TimeSformer-B, using CLIP-initialized weights pretrained on WebImageText (WIT)~\cite{radford2021clip} works slightly better than using those supervised pretrained on ImageNet-21k~\cite{deng2009imagenet,steiner2022train}.
Second, if we replace the visual encoder with a ViT-Base model as in CLIP, the performance drops by 1.5\% avg. mAP and 1.0\% avg. nDCG, indicating the necessity of using spatial-temporal visual encoder for learning video-language tasks.

\myparagraph{Effect of batch size}.
We study the effect of batch size of contrastive pre-training in \cref{tab:batch_size}.
The baseline method follows EgoVLP~\cite{lin2022egovlp} and uses a total batch size of 512.
We observe that the performance improves when increasing the batch size to 1,024.
The improvment diminishes if we further increase the batch size to 2,048.
Therefore, we use 1,024 as the default batch size to get our main results in \cref{sec:expt:main}.

\myparagraph{Effect of projection dimension}.
We compare different choices of the projection head's dimension in \cref{tab:projection_head}.
We can see that using 256 achieves the best performance compared to 128 or 512.

\myparagraph{Temperature in contrastive loss}.
In \cref{tab:temperature}, we study the effect of different temperatures in the contrastive loss (\cref{eq:loss}).
Note that we switch to a batch size of 1,024 based on the observation in \cref{tab:batch_size}.
We start with a learnable temperature of 0.07 following CLIP~\cite{radford2021clip}.
We can see that using a higher initial temperature $\tau_{n}$ for the pairs generated by \narrator achieves noticable gain over the one that uses the same initial temperature of 0.07 for both $\tau_{r}$ and $\tau_{n}$.
We found that the within-batch accuracy during contrastive training for \narrator's pairs is significantly higher than the one for \rephraser's pairs.
Our conjection is that the dual-encoders is more likely to overfit the \narrator's pairs. 
Therefore, we switch to a fixed temperature and find that using $\tau_{r}=\tau_{n}=0.07$ works better than all other settings, such as learnable temperature.

\begin{table}[t]
	\begin{center}
		\tablestyle{1pt}{1.05}
		\begin{tabular}{x{24}x{24}x{24}x{24}|x{42}x{42}}
			$\tau_{r}$ & learn & $\tau_{n}$ & learn & Avg. mAP & Avg. nDCG  \\
			\thickhline
			0.07 & \cmark & n/a & n/a & 25.6 & 28.8 \\
			0.07 & \cmark & 0.07 & \cmark & 25.7 & 29.0 \\
			0.07 & \cmark & 0.10 & \cmark & 26.8 & 29.6 \\
			0.07 & \cmark & 0.10 & \xmark & 27.4 & 29.8 \\
			0.07 & \xmark & n/a & n/a & 26.0& 29.0 \\
			0.07 & \xmark & 0.07 & \xmark & {\bf 29.5} & {\bf 31.1} \\
			0.07 & \xmark & 0.10 & \xmark & 27.4 & 29.8 \\
			\hline
		\end{tabular}
	\end{center}
	\caption{\textbf{Temperature in contrastive loss.} We observe that using a same fixed temperature for both \narrator's pairs and \rephraser's pairs works better than all other settings.}
	\label{tab:temperature}
\end{table}

\begin{figure}[t]
	\centering
	\includegraphics[width=\linewidth]{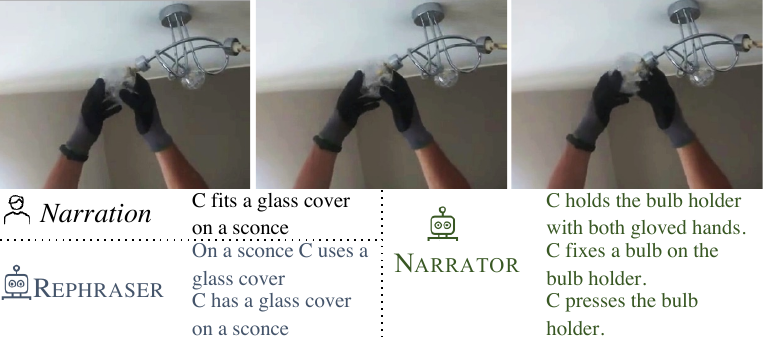}
	\includegraphics[width=\linewidth]{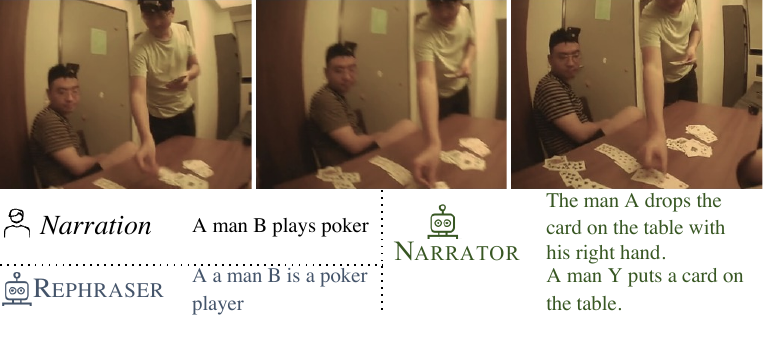}
	\includegraphics[width=\linewidth]{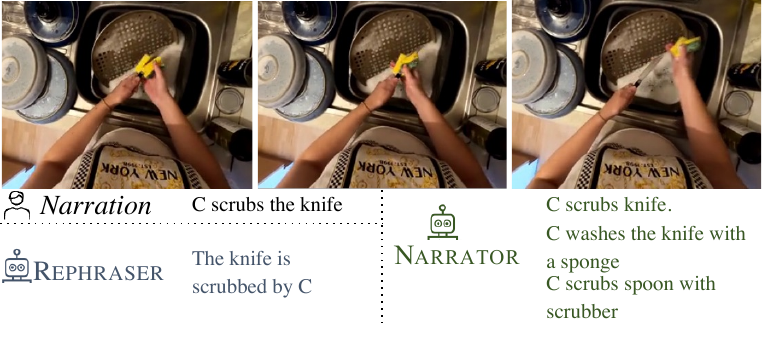}
	\includegraphics[width=\linewidth]{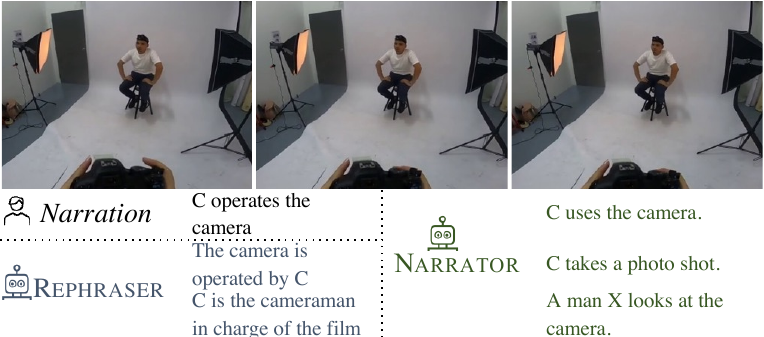}
	\includegraphics[width=\linewidth]{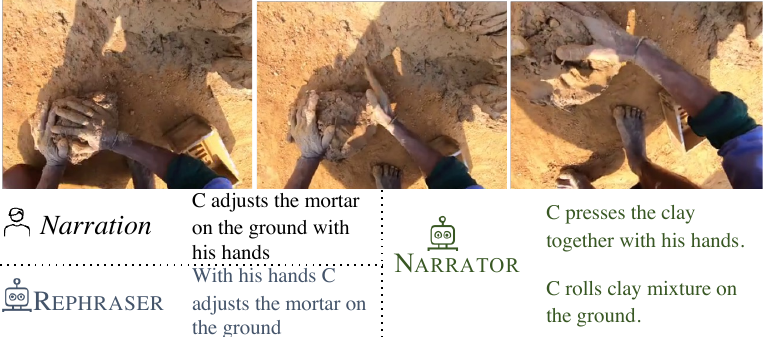}
	\caption{
		\textbf{More generated samples by our \narrator and \rephraser on Ego4D}.
		\narrator generates new descriptions of the action taking place, potentially focusing on other objects or person being interacted with.
		\rephraser not only changes the word order of the human narration but also diversifies it by using related verbs or nouns.
		Please refer to \cref{sec:appdx:qualitative} for discussion.
		\newline
	}
	\label{fig:appdx:qual_ego4d}
\end{figure}

\section{Qualitative Results}\label{sec:appdx:qualitative}
We provide more generated samples by our \narrator and \rephraser in \cref{fig:appdx:qual_ego4d}.
Note that our \narrator can generate reasonable captions from different views. For instance, \cref{fig:appdx:qual_ego4d}(d) illustrates that \narrator can describe the activities of both the camera wearer (starting with ``C'', which stands for ``Camera wearer'' in Ego4D) and the other person (starting with ``O'', which stands for ``Observer'' in Ego4D.

\section{Licenses}
The images in \cref{fig:teaser,fig:qual,fig:method,fig:appdx:qual_ego4d} are adapted from Ego4D videos.
The video id (\texttt{\$vid}) along with the start/end timestamp is provided below.
The video can be viewed via the url \url{https://visualize.ego4d-data.org/$vid} (License is required for access).
\begin{itemize}
	\item \cref{fig:teaser}: \\ \texttt{1bfac46e-f957-4495-9583-dbd7fa683225, 01:30:00-01:50:00}.
	\item \cref{fig:qual} (top): \\ \texttt{06919917-76bc-4adc-b944-2a722f165513, 00:00:08-00:00:10}.
	\item \cref{fig:qual} (bottom): \\ \texttt{cf7c12db-1a9e-46d3-96d6-38174bbe373c, 00:21:17-00:21:19}.
	\item \cref{fig:method}: \\ \texttt{3c0dffd0-e38e-4643-bc48-d513943dc20b, 00:00:12-00:00:14}.
	\item \cref{fig:appdx:qual_ego4d} (a): \\ \texttt{26054ab4-4967-47b5-9b6c-e8a62f9295e0, 00:08:09-00:08:10}.
	\item \cref{fig:appdx:qual_ego4d} (b): \\ \texttt{3130e00e-873a-4afb-93a6-7b07f3cf6597, 00:11:42-00:11:44}.
	\item \cref{fig:appdx:qual_ego4d} (c): \\ \texttt{def2e8dd-aaf7-467f-aa8f-46f654e6f4e0, 00:09:08-00:09:09}.
	\item \cref{fig:appdx:qual_ego4d} (d): \\ \texttt{ab865129-78fa-47d4-8a50-ff8c5533246f, 00:04:10-00:04:12}.
	\item \cref{fig:appdx:qual_ego4d} (e): \\ \texttt{58a01f3a-52ce-4024-ab3c-b179caf4dafd, 00:28:43-00:28:45}.
\end{itemize}

\end{document}